\documentclass[fullname]{clv2}
\pdfoutput=1
\issue{43}{1}{2017}

\usepackage{comment}
\usepackage{url}
\usepackage[utf8]{inputenc}
\usepackage{multirow}

\usepackage[normalem]{ulem}
\def\claim#1{\bgroup \uuline{[\emph{claim:} #1]}\egroup}
\def\premise#1{\bgroup \uline{[\emph{premise:} #1]}\egroup}
\def\backing#1{\bgroup \dotuline{[\emph{backing:} #1]}\egroup}
\def\rebuttal#1{\bgroup \uwave{[\emph{rebuttal: }#1]}\egroup}
\def\refutation#1{\bgroup \dashuline{[\emph{refutation:} #1]}\egroup}
\def\appealtoemotion#1{\bgroup \uline{[\emph{app-to-emot:} #1]}\egroup}
\def\tableex#1{\begin{scriptsize}\emph{$\bullet$ #1}\end{scriptsize}}

\newcommand{\textunderscript}[1]{$_{\mathrm{#1}}$}
\newcommand{\svmhmm}{SVM\textsuperscript{\emph{hmm}}\ }
\newcommand{\alphaunit}{\alpha_\mathrm{U}}

\usepackage[table]{xcolor}
\usepackage{pgfplotstable}

\pgfplotstableset{
    color cells/.style={
        col sep=comma,
        string type,
        postproc cell content/.code={%
                \pgfkeysalso{@cell content=\rule{0cm}{2.4ex}\cellcolor{black!##1}\pgfmathtruncatemacro\number{##1}\ifnum\number>50\color{white}\fi##1}%
                },
        columns/x/.style={
            column name={},
            postproc cell content/.code={}
        }
    }
}

\usepackage{caption}
\usepackage{subcaption}

\usepackage{lastpage}

\runningtitle{Argumentation Mining in User-Generated Web Discourse}

\runningauthor{Habernal and Gurevych}

\historydates{Submission received: 2 April 2015; revised version received: 20 April 2016; accepted for publication: 14 June 2016.}

\begin{document}

\title{Argumentation Mining in User-Generated Web Discourse}

\author{Ivan Habernal\thanks{Ubiquitous Knowledge Processing Lab (UKP-DIPF), German Institute for Educational Research, Schloßstraße 29, D-60486 Frankfurt am Main, Germany and  Technische Universit\"{a}t Darmstadt, Ubiquitous Knowledge Processing (UKP) Lab, TU Darmstadt - FB 20a Hochschulstrasse 10, D-64289 Darmstadt, Germany.}}
\affil{German Institute for Educational Research \\ and \\ Technische Universit\"{a}t Darmstadt}

\author{Iryna Gurevych\thanks{Technische Universit\"{a}t Darmstadt, Ubiquitous Knowledge Processing (UKP) Lab, TU Darmstadt - FB 20a Hochschulstrasse 10, D-64289 Darmstadt, Germany and Ubiquitous Knowledge Processing Lab (UKP-DIPF), German Institute for Educational Research, Schloßstraße 29, D-60486 Frankfurt am Main, Germany}}
\affil{Technische Universit\"{a}t Darmstadt \\ and \\ German Institute for Educational Research}

\maketitle

\begin{abstract}
The goal of argumentation mining, an evolving research field in computational linguistics, is to design methods capable of analyzing people's argumentation. In this article, we go beyond the state of the art in several ways. (i) We deal with actual Web data and take up the challenges given by the variety of registers, multiple domains, and unrestricted noisy user-generated Web discourse. (ii) We bridge the gap between normative argumentation theories and argumentation phenomena encountered in actual data by adapting an argumentation model tested in an extensive annotation study. (iii) We create a new gold standard corpus (90k tokens in 340 documents) and experiment with several machine learning methods to identify argument components. We offer the data, source codes, and annotation guidelines to the community under free licenses. Our findings show that argumentation mining in user-generated Web discourse is a feasible but challenging task.
\end{abstract}

\setcounter{page}{125}

\section{Introduction}
\label{sec:introduction}

The art of \emph{argumentation} has been studied since the early work of Aristotle, dating back to the 4th century BC \cite{Aristotle.1991}. It has been exhaustively examined from different perspectives, such as philosophy, psychology, communication studies, cognitive science, formal and informal logic, linguistics, computer science, educational research, and many others. In a recent and critically well-acclaimed study, \namecite{Mercier.Sperber.2011} even claim that argumentation is what drives humans to perform reasoning. From the pragmatic perspective, argumentation can be seen as a \emph{verbal activity oriented towards the realization of a goal} \cite{Micheli2011} or more in detail as \emph{a verbal, social, and rational activity aimed at convincing a reasonable critic of the acceptability of a standpoint by putting forward a constellation of one or more propositions to justify this standpoint} \cite{vanEemeren.et.al.2002}.

Analyzing argumentation from the computational linguistics point of view has very recently led to a new field called \emph{argumentation mining} \cite{Green.et.al.2014}. Despite the lack of an exact definition, researchers within this field usually focus on analyzing discourse on the pragmatics level and applying a certain argumentation theory to model and analyze textual data\footnote{Despite few recent multi-modal approaches to process argumentation and persuasion, e.g., \cite{Brilman.Scherer.2015a}, the main mode of argumentation mining is natural language text.} at hand.

Our motivation for argumentation mining stems from a practical \emph{information seeking} perspective from the user-generated content on the Web. For example, when users search for information in user-generated Web content to facilitate their personal decision making related to controversial topics, they lack tools to overcome the current information overload. One particular use-case example dealing with a forum post discussing \emph{private versus public schools} is shown in Figure \ref{fig:motivation.gist}. Here, the lengthy text on the left-hand side is transformed into an argument gist on the right-hand side by (i) analyzing argument components and (ii) summarizing their content. Figure \ref{fig:motivation.reasons} shows another use-case example, in which users search for reasons that underpin certain standpoint in a given controversy (which is \emph{homeschooling} in this case). In general, the output of automatic argument analysis performed on the large scale in Web data can provide users with analyzed arguments to a given topic of interest, find the evidence for the given controversial standpoint, or help to reveal flaws in argumentation of others.

\begin{figure}
\begin{minipage}[t]{0.48\textwidth}
\textbf{Original text} \\
\begin{scriptsize}
The public schooling system is not as bad as some may think. Some mentioned that those who are educated in the public schools are less educated, well I actually think it would be in the reverse. Student who study in the private sector actually pay a fair amount of fees to do so and I believe that the students actually get let off for a lot more than anyone would in a public school. And its all because of the money. In a private school, a student being expelled or suspended is not just one student out the door, its the rest of that students schooling life fees gone. Whereas in a public school, its just the student gone. I have always gone to public schools and when I finished I got into University. I do not feel disadvantaged at all.
\end{scriptsize}
\end{minipage}
\hfill
\begin{minipage}[t]{0.48\textwidth}
\textbf{Extracted argument gist} \\
\textbf{I claim that} public schools are good \textbf{because} students in private schools are only source of money. \textbf{I can back-up my argument:} I have always gone to public schools and when I finished I got into University. I do not feel disadvantaged at all. \textbf{On the other hand} some mentioned that those who are educated in the public schools are less educated.
\end{minipage}
\caption{\label{fig:motivation.gist} Motivation example 1: Extracting argument gist by means of analyzing the argument structure and summarizing the argument components. The bold phrases are generated automatically and invoke the component function;
\protect\namecite{Madnani.et.al.2012} refers to these organizational elements as \emph{shells}. \emph{Doc\#4733, forum post, public-private schools}.}
\end{figure}

\begin{figure}
\begin{small}
\begin{minipage}[t]{0.48\textwidth}
\noindent \textbf{Reasons for homeschooling} \\
$\bullet$ Schools provide a totally unstimulating environment. \\
$\bullet$ Lesson plans (and the national curriculum) are the death of real education. \\
$\bullet$ Evidence including our own suggests strongly that this kind of education prepares children to enter further and higher education, or the workforce - and offers them the freedom to learn in the ways that suit them best. \\
$\bullet$ We teach our children how to learn, not merely how to pass tests.
\end{minipage}
\hfill
\begin{minipage}[t]{0.48\textwidth}
\textbf{Reasons against homeschooling} \\
$\bullet$ Keeping your kids away from knowledge that you don’t like is a moral crime. \\
$\bullet$ Religious zealotry is no excuse for raising a kid devoid of a proper education. \\
$\bullet$ Consciously depriving a child of an adequate education solely because ”father knows best,” or thinks he does, is tantamount to child abuse.
\end{minipage}
\end{small}
\caption{\label{fig:motivation.reasons} Motivation example 2: Extracting evidence for a certain standpoint with respect to a given controversial topic. All statements are taken from the corpus introduced in this article.}
\end{figure}

Satisfying the above-mentioned information needs cannot be directly tackled by current methods for, e.g., opinion mining, questions answering,\footnote{These research fields are still related and complementary to argumentation mining. For example, personal decision-making queries (such as \emph{``Should I homeschool my children?''}) might be tackled by researches exploiting social question-answering sites.} or summarization\footnote{The role of argumentation moves in summarizing scientific articles was examined by  \namecite{Teufel2002}.} and requires novel approaches within the argumentation mining field.
Although user-generated Web content has already been considered in argumentation mining, many limitations and research gaps can be identified in the existing works.
First, the scope of the current approaches is restricted to a particular domain or register, e.g., hotel reviews \cite{Wachsmuth.et.al.2014b}, Tweets related to local riot events \cite{Llewellyn2014}, student essays \cite{Stab.Gurevych.2014}, airline passenger rights and consumer protection \cite{Park.Cardie.2014}, or renewable energy sources \cite{Goudas.et.al.2014}.
Second, not all the related works are tightly connected to argumentation theories, resulting into a gap between the substantial research in argumentation itself and its adaptation in NLP applications.
Third, as an emerging research area, argumentation mining still suffers from a lack of labeled corpora, which is crucial for designing, training, and evaluating the algorithms. Although some works have dealt with creating new data sets, the reliability (in terms of inter-annotator agreement) of the annotated resources is often unknown \cite{Feng.Hirst.2011,Mochales2011,Walton.2012,FlorouKonstantopoulos2013,Villalba.et.al.2012}.

Annotating and automatically analyzing arguments in unconstrained user-generated Web discourse represent challenging tasks.
So far, the research in argumentation mining ``has been conducted on domains like news articles, parliamentary records and legal documents, where the documents contain well-formed explicit arguments, i.e., propositions with supporting reasons and evidence present in the text'' \cite[p.~29]{Park.Cardie.2014}. \namecite[p.~50]{Boltuzic.Snajder.2014} point out that ``unlike in debates or other more formal argumentation sources, the arguments provided by the users, if any, are less formal, ambiguous, vague, implicit, or often simply poorly worded.'' 
Another challenge stems from the different nature of argumentation theories and computational linguistics. Whereas computational linguistics is mainly descriptive, the empirical research that is carried out in argumentation theories does not constitute a test of the theoretical model that is favored, because the model of argumentation is a \emph{normative} instrument for assessing the argumentation \cite[pp.~11]{vanEmeren.et.al.2014}. So far, no fully fledged descriptive argumentation theory based on empirical research has been developed, thus feasibility of adapting argumentation models to the Web discourse represents an open issue.

These challenges can be formulated into the following research questions:

\begin{itemize}
\item Can we adapt models from argumentation theories, that have been usually lacking empirical evidence on large real-world corpora, for modeling argumentation in user-generated Web content?
\item What are the desired properties of the argumentation model and is there a trade-off between model complexity and annotation reliability?
\item What phenomena are typical of argumentation on the Web, how to approach their modeling, and what challenges do they pose?
\item What is the impact of different controversial topics and are there differences in argumentation between various registers?
\item What computational approaches can be used to analyze arguments on the Web?
\end{itemize}

In this article, we push the boundaries of the argumentation mining field by focusing on several novel aspects. We tackle the above-mentioned research questions as well as the previously discussed challenges and issues. First, we target user-generated Web discourse from several domains across various registers, to examine how argumentation is communicated in different contexts. Second, we bridge the gap between argumentation theories and argumentation mining through selecting the argumenation model based on research into argumentation theories and related fields in communication studies or psychology. In particular, we adapt normative models from argumentation theory to perform empirical research in NLP and support our application of argumentation theories with an in-depth reliability study. Finally, we use state-of-the-art NLP techniques in order to build robust computational models for analyzing arguments that are capable of dealing with a variety of genres on the Web.\footnote{We used the dataset and core methods from this article in our sequel publication \cite{Habernal.Gurevych.2015}. The main difference is that the this article focuses mainly on corpus annotation, analysis, and argumentation on Web in general, while in \cite{Habernal.Gurevych.2015} we explored whether methods for recognizing argument components benefit from using semi-supervised features obtained from noisy debate portals.}

\subsection{Our contributions}

We create a \textbf{new corpus} which is, to the best of our knowledge, the largest corpus that has been annotated within the argumentation mining field to date. We choose several target domains from educational controversies, such as \emph{homeschooling}, \emph{single-sex education}, or \emph{mainstreaming}.\footnote{Controversial educational topics attract a wide range of participants, such as parents, journalists, education experts, policy makers, or students, which contributes to the linguistic breadth of the discourse.} A novel aspect of the corpus is its coverage of different registers of \textbf{user-generated Web content}, such as comments to articles, discussion forum posts, blog posts, as well as professional newswire articles.

Since the data come from a variety of sources and no assumptions about its actual content with respect to argumentation can be drawn, we conduct two extensive \textbf{annotation studies}. In the first study, we tackle the problem of relatively high ``noise'' in the retrieved data. In particular, not all of the documents are related to the given topics in a way that makes them candidates for further deep analysis of argumentation (this study results into 990 annotated documents). In the second study, we discuss the selection of an appropriate argumentation model based on evidence in argumentation research and propose a model that is suitable for analyzing micro-level argumention in user-generated Web content. Using this model, we annotate 340 documents (approx. 90,000 tokens), reaching a substantial inter-annotator agreement. We provide a hand-analysis of all the phenomena typical to argumentation that are prevalent in our data. These findings may also serve as empirical evidence to issues that are on the spot of current argumentation research. 

From the computational perspective, we experiment on the annotated data using various machine learning methods in order to extract argument structure from documents. We propose several novel \textbf{feature sets} and identify configurations that run best in in-domain and cross-domain scenarios. To foster research in the community, \textbf{we provide the annotated data} as well as all the experimental software under free license.\footnote{\url{https://www.ukp.tu-darmstadt.de/data/argumentation-mining/}}

\paragraph{Outline}

The rest of the article is structured as follows. First, we provide an essential background in argumentation theory in section \ref{sec:theoretial.background}. Section \ref{sec:related.work} surveys related work in several areas. Then we introduce the dataset and two annotation studies in section \ref{sec:annotation.studies}. Section \ref{sec:experiments} presents our experimental work and discusses the results and errors and section \ref{sec:conclusion} concludes this article.

\section{Theoretical background}
\label{sec:theoretial.background}

Let us first present some definitions of the term \textbf{argumentation} itself. \namecite[p.~3]{Ketcham.1917} defines argumentation as \emph{``the art of persuading others to think or act in a definite way. It includes all writing and speaking which is persuasive in form.''} According to \namecite{MacEwan.1898}, \emph{``argumentation is the process of proving or disproving a proposition. Its purpose is to induce a new belief, to establish truth or combat error in the mind of another.''} \namecite[p.~2]{Freeley.Steinberg.2008} narrow the scope of argumentation to \emph{``reason giving in communicative situations by people whose purpose is the justification of acts, beliefs, attitudes, and values.''} Although these definitions vary, the purpose of argumentation remains the same -- to persuade others.

We would like to stress that our perception of argumentation goes beyond somehow limited \emph{giving reasons} \cite{Freeley.Steinberg.2008,Damer.2013}. Rather, we see the goal of argumentation as to persuade \cite{Ketcham.1917,Nettel2011,Mercier.Sperber.2011}. \textbf{Persuasion} can be defined as \textit{a successful intentional effort at influencing another's mental state through communication in a circumstance in which the persuadee has some measure of freedom} \cite[p.~5]{OKeefe.2002}, although, as \namecite{OKeefe2011} points out, there is no correct or universally-endorsed definition of either `persuasion' or `argumentation'. However, broader understanding of argumentation as a \emph{means} of persuasion allows us to take into account not only reasoned discourse, but also non-reasoned mechanisms of influence, such as emotional appeals \cite{Blair2011}.

Having an \textbf{argument} as a product within the argumentation process, we should now define it. One typical definition is that \emph{an argument is a claim supported by reasons} \cite[p.~6]{Schiappa.Nordin.2013}. The term \textbf{claim} has been used since 1950's, introduced by \namecite{Toulmin.1958}, and in argumentation theory it is a synonym for \emph{standpoint} or \emph{point of view}. It refers to what is an issue in the sense what is being argued about. The presence of a standpoint is thus crucial for argumentation analysis. However, the claim as well as other parts of the argument might be implicit; this is known as \emph{enthymematic argumentation}, which is rather usual in ordinary argumentative discourse \cite{Amossy2009}.

One fundamental problem with the definition and formal description of arguments and argumentation is that there is no agreement even among argumentation theorists. As \namecite[p.~29]{vanEmeren.et.al.2014} admit in their very recent and exhaustive survey of the field, \emph{''as yet, there is no unitary theory of argumentation that encompasses the logical, dialectical, and rhetorical dimensions of argumentation and is universally accepted. The current state of the art in argumentation theory is characterized by the coexistence of a variety of theoretical perspectives and approaches, which differ considerably from each other in conceptualization, scope, and theoretical refinement.''} 

\subsection{Argumentation models}

Despite the missing consensus on the ultimate argumentation theory, various argumentation models have been proposed that capture argumentation on different levels.
Argumentation models abstract from the language level to a concept level that stresses the links between the different components of an argument or how arguments relate to each other \cite{Prakken.Vreeswijk.2002}.
\namecite{Bentahar.et.al.2010} propose a taxonomy of argumentation models, that is horizontally divided into three categories -- \textbf{micro-level} models, \textbf{macro-level} models, and \textbf{rhetorical} models.

In this article, we deal with argumentation on the \textbf{micro-level} (also called argumentation as a product or monological models). Micro-level argumentation focuses on the structure of a single argument. By contrast, \textbf{macro-level} models (also called \emph{dialogical} models) and \textbf{rhetorical} models highlight the process of argumentation in a dialogue \cite[p.~215]{Bentahar.et.al.2010}. In other words, we examine the structure of a single argument produced by a single author in term of its components, not the relations that can exist among arguments and their authors in time. A detailed discussion of these different perspectives can be found, e.g., in \cite{Blair.2004,Johnson.2000,Reed.Walton.2003,Micheli2011,OKeefe.1982,Rapanta.et.al.2013}.\footnote{There are, however, some argumentation theorist who disagree with this distinction and consider argumentation purely as dialogical. \namecite{Freeman.2011} sees the argument as a process that is implicitly present even if the argumentation is a written text, which others treat as argument as product. For a deep discussion of opposing views on dialectical nature of argumentation, we would point to \cite[p.~53]{Freeman.2011}, \namecite{Finocchiaro.2005} or to the pragma-dialectical approach by \namecite{Eemeren.Grootendorst.1984}.}

\subsection{Dimensions of argument}
\label{sec:dimensions.of.argument}

The above-mentioned models focus basically only on one dimension of the argument, namely the \textbf{logos} dimension. According to the classical Aristotle's theory \cite{Aristotle.1991}, argument can exist in three dimensions, which are logos, pathos, and ethos. \textbf{Logos} dimension represents a proof by reason, an attempt to persuade by establishing a logical argument. For example, syllogism belongs to this argumentation dimension \cite{Rapp2012,Amossy2009}. \textbf{Pathos} dimension makes use of appealing to emotions of the receiver and impacts its cognition \cite{Micheli2008}. \textbf{Ethos} dimension of argument relies on the credibility of the arguer. This distinction will have practical impact later in section \ref{sec:annotation.study.2} which deals with argumentation on the Web.

\subsection{Original Toulmin's model}
\label{sec:toulmin.orig}

We conclude the theoretical section by presenting one (micro-level) argumentation model in detail -- a widely used conceptual model of argumentation introduced by \namecite{Toulmin.1958}, which we will henceforth denote as the \emph{Toulmin's original model}.\footnote{Henceforth, we will refer to the updated edition of \cite{Toulmin.1958}, namely \cite{Toulmin.2003}.} This model will play an important role later in the annotation studies (section \ref{sec:annotation.study.2}) and experimental work (section \ref{sec:identification.of.argument.components}). The model consists of six parts, referred as \textbf{argument components}, where each component plays a distinct role.

\begin{description}
\item[Claim] is an assertion put forward publicly for general acceptance \cite[p.~29]{Toulmin.et.al.1984} or the conclusion we seek to establish by our arguments \cite[p.~153]{Freeley.Steinberg.2008}.
\item[Data (Grounds)] It is the evidence to establish the foundation of the claim \cite{Schiappa.Nordin.2013} or, as simply put by Toulmin, \emph{``the data represent what we have to go on.''} \cite[p.~90]{Toulmin.2003}. The name of this concept was later changed to \emph{grounds} in \cite{Toulmin.et.al.1984}.
\item[Warrant] The role of \emph{warrant} is to justify a logical inference from the \emph{grounds} to the \emph{claim}.
\item[Backing] is a set of information that stands behind the \emph{warrant}, it assures its trustworthiness.
\item[Qualifier] limits the degree of certainty under which the argument should be accepted. It is the degree of force which the \emph{grounds} confer on the \emph{claim} in virtue of the \emph{warrant} \cite[p.~93]{Toulmin.2003}.
\item[Rebuttal] presents a situation in which the \emph{claim} might be defeated.
\end{description}

\begin{figure}
	\centering
	\includegraphics[width=0.5\linewidth]{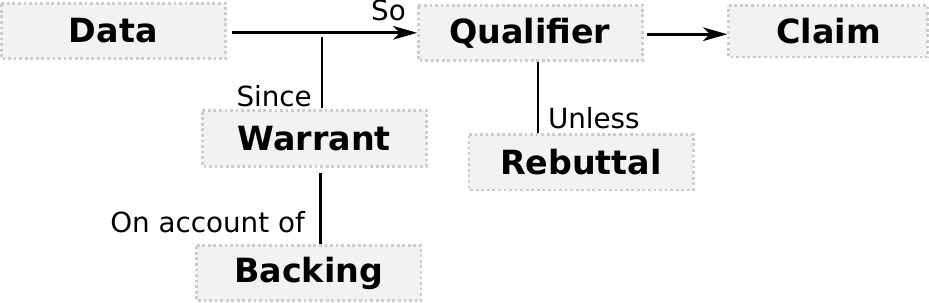}
	\caption{\label{fig:toulmin-orig}Original Toulmin's model of argument.}
\end{figure}

\begin{figure}
{[}Harry was born in Bermuda.{]}\textunderscript{Data}
\textbf{Since} {[}A man born in Bermuda will generally be a British subject.{]}\textunderscript{Warrant}
\textbf{On account of} {[}The following statuses and other legal provisions: (...){]}\textunderscript{Backing}
\textbf{So,} {[}presumably{]}\textunderscript{Qualifier}
\textbf{Unless} {[}Both his parents were aliens{]}\textunderscript{Rebuttal}
{[}Harry is a British subject.{]}\textunderscript{Claim}
\caption{\label{fig:toulmin.orig.example.text} Example of an argument using Toulmin's model \protect\cite{Toulmin.2003}.}
\end{figure}

A schema of the Toulmin's original model is shown in Figure \ref{fig:toulmin-orig}. The lines and arrows  symbolize implicit relations between the components. An example of an argument rendered using the Toulmin's scheme can be seen in Figure \ref{fig:toulmin.orig.example.text}.

We believe that this theoretical overview should provide sufficient background for the argumentation mining research covered in this article; for further references, we recommend for example \cite{vanEmeren.et.al.2014}.

\section{Related work in computational linguistics}
\label{sec:related.work}

We structure the related work into three sub-categories, namely \emph{argumentation mining}, \emph{stance detection}, and \emph{persuasion and on-line dialogs}, as these areas are closest to this article's focus. For a recent overview of general discourse analysis see \cite{Webber.et.al.2011}. Apart from these, research on computer-supported argumentation has been also very active; see, e.g., \cite{Scheuer.et.al.2013} for a survey of various models and argumentation formalisms from the educational perspective or \cite{Schneider.et.al.2013} which examines argumentation in the Semantic Web.

\subsection{Argumentation Mining}
\label{sec:rel.work.arg.min}

The argumentation mining field has been evolving very rapidly in the recent years, resulting into several workshops co-located with major NLP conferences. 
We first present related works with a focus on annotations and then review experiments with classifying argument components, schemes, or relations.

\subsubsection{Annotation studies}

One of the first papers dealing with annotating argumentative discourse was Argumentative Zoning for scientific publications \cite{Teufel.et.al.1999}. Later, \namecite{Teufel.et.al.2009} extended the original 7 categories to 15 and annotated 39 articles from two domains, where each sentence is assigned a category. The obtained Fleiss' $\kappa$ was 0.71 and 0.65. In their approach, they tried to deliberately ignore the domain knowledge and rely only on general, rhetorical and logical aspect of the annotated texts. By contrast to our work, argumentative zoning is specific to scientific publications and has been developed solely for that task.

\namecite{Reed.Rowe.2004} presented \emph{Araucaria}, a tool for argumentation diagramming which supports both convergent and linked arguments, missing premises (enthymemes), and refutations. They also released the \emph{AracuariaDB} corpus which has later been used for experiments in the argumentation mining field. However, the creation of the dataset in terms of annotation guidelines and reliability is not reported -- these limitations as well as its rather small size have been identified \cite{Feng.Hirst.2011}.

\namecite{Biran.Rambow.2011} identified justifications for subjective claims in blog threads and Wikipedia talk pages. The data were annotated with claims and their justifications reaching $\kappa$ 0.69, but a detailed description of the annotation approach was missing.

\namecite[p.~1078]{Schneider.et.al.2013b} annotated Wikipedia talk pages about deletion using 17 Walton's schemes \cite{Walton.2007}, reaching a moderate agreement (Cohen's $\kappa$ 0.48) and concluded that their analysis technique can be reused, although ``\emph{it is intensive and difficult to apply}.''

\namecite{Stab.Gurevych.2014} annotated 90 argumentative essays (about 30k tokens), annotating claims, major claims, and premises and their relations (support, attack). They reached Krippendorff's $\alphaunit$ 0.72 for argument components and Krippendorff's $\alpha$ 0.81 for relations between components.

\namecite{Rosenthal2012} annotated sentences that are opinionated claims, in which the author expresses a belief that should be adopted by others. Two annotators labeled sentences as claims without any context and achieved Cohen's $\kappa$ 0.50 (2,000 sentences from LiveJournal) and 0.56 (2,000 sentences from Wikipedia).

\namecite{Aharoni.et.al.2014} performed an annotation study in order to find context-dependent claims and three types of context-dependent evidence in Wikipedia, that were related to 33 controversial topics. The claim and evidence were annotated in 104 articles. The average Cohen's $\kappa$ between a group of 20 expert annotators was 0.40. Compared to our work, the linguistic properties of Wikipedia are qualitatively different from other user-generated content, such as blogs or user comments \cite{Ferschke.2014}.

\namecite{Wacholder.et.al.2014} annotated ``argument discourse units'' in blog posts and criticized the Krippendorff's $\alphaunit$ measure. They proposed a new inter-annotator metric by taking the most overlapping part of one annotation as the ``core'' and all annotations as a ``cluster''. The data were extended by \namecite{Ghosh2014}, who annotated ``targets'' and ``callouts'' on the top of the units.

\namecite{Park.Cardie.2014} annotated about 10k sentences from 1,047 documents into four types of argument propositions with Cohen's $\kappa$ 0.73 on 30\% of the dataset. Only 7\% of the sentences were found to be non-argumentative.

\namecite{Faulkner2014} used Amazon Mechanical Turk to annotate 8,179 sentences from student essays. Three annotators decided whether the given sentence offered reasons for or against the main prompt of the essay (or no reason at all; 66\% of the sentences were found to be neutral and easy to identify). The achieved Cohen's $\kappa$ was 0.70.

The research has also been active on non-English datasets. \namecite{Goudas.et.al.2014} focused on user-generated Greek texts. They selected 204 documents and manually annotated sentences that contained an argument (760 out of 16,000). They distinguished claims and premises, but the claims were always implicit. However, the annotation agreement was not reported, neither was the number of annotators or the guidelines.
A study on annotation of arguments was conducted by \namecite{Peldszus.Stede.2013}, who evaluate agreement among 26 ``naive" annotators (annotators with very little training). They manually constructed 23 German short texts, each of them contains exactly one central claim, two premises, and one objection (rebuttal or undercut) and analyzed annotator agreement on this artificial data set. \namecite{Peldszus.2014} later achieved higher inter-rater agreement with expert annotators on an extended version of the same data.  \namecite{Kluge.2014} built a corpus of argumentative German Web documents, containing 79 documents from 7 educational topics, which were annotated by 3 annotators according to the claim-premise argumentation model. The corpus comprises 70,000 tokens and the inter-annotator agreement was 0.40 (Krippendorff's $\alphaunit$).  \namecite{Houy.et.al.2013} targeted argumentation mining of German legal cases.

Table \ref{tab:related-work} gives an overview of annotation studies with their respective argumentation model, domain, size, and agreement. It also contains other studies outside of computational linguistics and few proposals and position papers.

\begin{table*}
\begin{scriptsize}
\begin{tabular}{p{2cm}|p{2cm}|p{2.5cm}|p{2.2cm}|p{2.0cm}}
\textbf{Source} & \textbf{Arg. Model} & \textbf{Domain} & \textbf{Size} & \textbf{IAA} \\ \hline
\namecite{Newman1991} & \namecite{Toulmin.1958} &  legal domain (People vs. Carney, U.S. Supreme Court) & qualitative & N/A \\ \hline
\namecite{Bal2010} & proprietary & socio-political newspaper editorials & 56 documents & Cohen's $\kappa$ \newline (0.80) \\ \hline
\namecite{Feng.Hirst.2011}  & \namecite{Walton.et.al.2008} \newline (top 5 schemes) & legal domain (AracuariaDB corpus, 61\% subset annotated with Walton scheme) & $\approx$ 400 arguments  & not reported \newline claimed to be small \\ \hline
\namecite{Biran.Rambow.2011} & proprietary & Wikipedia Talk pages, blogs & 309 + 118 & Cohen's $\kappa$ \newline (0.69) \\ \hline
\namecite{Georgila.et.al.2011}  & proprietary & general discussions (negotiations between florists)  & 21 dialogs & Krippendorff's $\alpha$ \newline (0.37-0.56) \\ \hline
\namecite{Mochales2011}  & Claim-Premise based on \namecite{Freeman1991} & legal domain (AracuariaDB corpus, European Human Rights Council) & 641 documents w/ 641 arguments (AracuariaDB) \newline 67 documents w/ 257 arguments (EHRC) & not reported \\ \hline
\namecite{Walton.2012}  & \namecite{Walton.et.al.2008} \newline (14 schemes) & political argumentation & 256 arguments & not reported \\ \hline
\namecite{Rosenthal2012}  & opinionated claim, sentence level &  blog posts, Wikipedia discussions & 4000 sentences & Cohen's $\kappa$ \newline (0.50-0.57) \\ \hline
\namecite{Conrad2012}  & proprietary \newline (spans of arguing subjectivity) & editorials and blog post about Obama Care & 84 documents & Cohen's $\kappa$ \newline (0.68) \newline on 10 documents \\ \hline
\namecite{Schneider2012} & proprietary, argumentation schemes & camera reviews & N/A \newline (proposal/position paper) & N/A \\ \hline
\namecite{Schneider2012a} & \namecite{Dung.1995} + \namecite{Walton.et.al.2008} & unspecified social media & N/A \newline (proposal/position paper) & N/A \\ \hline
\namecite{Villalba.et.al.2012} & proprietary, RST & hotel reviews, hi-fi products, political campaign & 50 documents	 & not reported \\ \hline
\namecite{Peldszus2013a} & \namecite{Freeman1991} + RST & Potsdam Commentary Corpus & N/A \newline (proposal/position paper) & N/A \\ \hline
\namecite{FlorouKonstantopoulos2013} & none & public policy making & 69 argumentative segments / 322 non-argumentative segments & not reported \\ \hline
\namecite{Peldszus.Stede.2013} & based on \namecite{Freeman1991} & not reported, artificial documents created for the study & 23 short documents & Fleiss' $\kappa$ \newline multiple results \\ \hline
\namecite{Sergeant2013} & N/A & Car Review Corpus (CRC) & N/A \newline (proposal/position paper) & N/A \\ \hline
\namecite{Wachsmuth.et.al.2014b} & none & hotel reviews & 2100 reviews & Fleiss' $\kappa$ \newline (0.67) \\ \hline
\namecite{Procter.et.al.2013} & proprietary \newline (Claim, Counter-claim) & Riot Twitter Corpus & 7729 tweets under `Rumours' category & percentage agreement \newline (89\% -- 96\%)\\ \hline
\namecite{Stab.Gurevych.2014} & Claim-Premise based on \namecite{Freeman1991} & student essays & 90 documents & Kripp. $\alpha_U$ (0.72) \newline Kripp. $\alpha$ (0.81) \\ \hline
\namecite{Aharoni.et.al.2014} & proprietary (claims, evidence) & Wikipedia & 104 documents & Cohen's $\kappa$ \newline (0.40) \\ \hline
\namecite{Park.Cardie.2014} & proprietary (argument propositions) & policy making (passenger rights and consumer protection) & 1047 documents & Cohen's $\kappa$ \newline (0.73) \\ \hline
\namecite{Goudas.et.al.2014} & proprietary (premises) & social media & 204 documents & not reported \\ \hline
\namecite{Faulkner2014} & none (``supporting argument'') & student essays & 8176 sentences & Cohen's $\kappa$ \newline (0.70)
\end{tabular} 
\end{scriptsize}
\caption{Previous works on annotating argumentation. IAA = Inter-annotator agreement; N/A = not applicable.}
\label{tab:related-work}
\end{table*}

\subsubsection{Argument analysis}

Arguments in the legal domain were targeted in \cite{Mochales2011}. Using argumentation formalism inspired by \namecite{Walton.2012}, they employed multinomial Naive Bayes classifier and maximum entropy model for classifying argumentative sentences on the \emph{AraucariaDB} corpus \cite{Reed.Rowe.2004}. The same test dataset was used by \namecite{Feng.Hirst.2011}, who utilized the C4.5 decision classifier. \namecite{Rooney.et.al.2012} investigated the use of convolution kernel methods for classifying whether a sentence belongs to an argumentative element or not using the same corpus.

\namecite{Stab.Gurevych.2014b} classified sentences to four categories (none, major claim, claim, premise) using their previously annotated corpus \cite{Stab.Gurevych.2014} and reached 0.72 macro-$F_1$ score. In contrast to our work, their documents are expected to comply with a certain structure of argumentative essays and are assumed to always contain argumentation.

\namecite{Biran.Rambow.2011} identified justifications on the sentence level using a naive Bayes classifier over a feature set based on statistics from the RST Treebank, namely n-grams which were manually processed by deleting n-grams that ``\emph{seemed irrelevant, ambiguous or domain-specific}.''

\namecite{Llewellyn2014} experimented with classifying tweets into several argumentative categories, namely claims and counter-claims (with and without evidence) and verification inquiries previously annotated by \namecite{Procter.et.al.2013}. They used unigrams, punctuations, and POS as features in three classifiers.

\namecite{Park.Cardie.2014} classified propositions into three classes 
(unverifiable, verifiable non-experimental, and verifiable experimental) and ignored non-argumentative texts. Using multi-class SVM and a wide range of features (n-grams, POS, sentiment clue words, tense, person) they achieved Macro$F_1$ 0.69.

\namecite{Peldszus.2014} experimented with a rather complex labeling schema of argument segments, but their data were artificially created for their task and manually cleaned, such as removing segments that did not meet the criteria or non-argumentative segments.

In the first step of their two-phase approach, \namecite{Goudas.et.al.2014} sampled the dataset to be balanced and identified argumentative sentences with $F_1$ 0.77 using the maximum entropy classifier. For identifying premises, they used BIO encoding of tokens and achieved $F_1$ score 0.42 using CRFs.

\namecite{Saint-Dizier.2012} developed a Prolog engine using a lexicon of 1300 words and a set of 78 hand-crafted rules with the focus on a particular argument structure ``reasons supporting conclusions'' in French.

Taking the dialogical perspective, \namecite{Cabrio.Villata.2012} built upon an argumentation framework proposed by \namecite{Dung.1995} which models arguments within a graph structure and provides a reasoning mechanism for resolving accepted arguments. For identifying support and attack, they relied on existing research on textual entailment \cite{Dagan.et.al.2009}, namely using the off-the-shelf \emph{EDITS} system. The test data were taken from a debate portal \emph{Debatepedia} and covered 19 topics. Evaluation was performed in terms of measuring the acceptance of the ``main argument" using the automatically recognized entailments, yielding $F_1$ score about 0.75. By contrast to our work which deals with micro-level argumentation, the Dung's model is an abstract framework intended to model dialogical argumentation.

Finding a bridge between existing discourse research and argumentation has been targeted by several researchers. \namecite{Peldszus2013a} surveyed literature on argumentation and proposed utilization of Rhetorical Structure Theory (RST) \cite{Mann.Thompson.1987}. They claimed that RST is by its design well-suited for studying argumentative texts, but an empirical evidence has not yet been provided. Penn Discourse Tree Bank (PDTB) \cite{Prasad2008} relations have been under examination by argumentation mining researchers too. \namecite{Cabrio2013b} examined a connection between five Walton's schemes and discourse markers in PDTB, however an empirical evaluation is missing.

\subsection{Stance detection}

Research related to argumentation mining also involves \emph{stance detection}. In this case, the whole document (discussion post, article) is assumed to represent the writer's standpoint to the discussed topic. Since the topic is stated as a controversial question, the author is either \emph{for} or \emph{against} it.

\namecite{Somasundaran.Wiebe.2009} built a computational model for recognizing stances in dual-topic debates about named entities in the electronic products domain by combining preferences learned from the Web data and discourse markers from PDTB \cite{Prasad2008}.
\namecite{Hasan.Ng.2013} determined stance in on-line ideological debates on four topics using data from createdebate.com, employing supervised machine learning and features ranging from n-grams to semantic frames.
Predicting stance of posts in Debatepedia as well as external articles using a probabilistic graphical model was presented in \cite{Gottipati.et.al.2013}. This approach also employed sentiment lexicons and Named Entity Recognition as a preprocessing step and achieved accuracy about 0.80 in binary prediction of stances in debate posts.

Recent research has involved joint modeling, taking into account information about the users, the dialog sequences, and others.
\namecite{Hasan.Ng.2012} proposed machine learning approach to debate stance classification by leveraging contextual information and author's stances towards the topic.
\namecite{Qiu.et.al.2013} introduced a computational debate side model to cluster posts or users by sides for general threaded discussions using a generative graphical model employing words from various subjectivity lexicons as well as all adjectives and adverbs in the posts.
\namecite{Qiu.Jiang.2013} proposed a graphical model for viewpoint discovery in discussion threads.
\namecite{Burfoot.et.al.2011} exploited the informal citation structure in U.S. Congressional floor-debate transcripts and use a collective classification which outperforms methods that consider documents in isolation.

Some works also utilize argumentation-motivated features.
\namecite{Park.et.al.2011} dealt with contentious issues in Korean newswire discourse. Although they annotate the documents with ``argument frames'', the formalism remains unexplained and does not refer to any existing research in argumentation.
\namecite{Walker.et.al.2012b} incorporated features with some limited aspects of the argument structure, such as cue words signaling rhetorical relations between posts, POS generalized dependencies, and a representation of the parent post (context) to improve stance classification over 14 topics from convinceme.net.

\subsection{Online persuasion}

Another stream of research has been devoted to persuasion in online media, which we consider as a more general research topic than argumentation.

\namecite{Schlosser.2011} investigated persuasiveness of online reviews and concluded that presenting two sides is not always more helpful and can even be less persuasive than presenting one side.
\namecite{Mohammadi.et.al.2013} explored persuasiveness of speakers in YouTube videos and concluded that people are perceived more persuasive in video than in audio and text.
\namecite{Miceli.et.al.2006} proposed a computational model that attempts to integrate emotional and non-emotional persuasion.
In the study of \namecite{Murphy.2001}, persuasiveness was assigned to 21 articles (out of 100 manually preselected) and four of them are later analyzed in detail for comparing the perception of persuasion between expert and students.
\namecite{Bernard.et.al.2012} experimented with children's perception of discourse connectives (namely with ``because'') to link statements in arguments and found out that 4- and 5-years-old and adults are sensitive to the connectives.
\namecite{Le.2004} presented a study of persuasive texts and argumentation in newspaper editorials in French.

A coarse-grained view on dialogs in social media was examined by \namecite{Bracewell.et.al.2013}, who proposed a set of 15 social acts (such as agreement, disagreement, or supportive behavior) to infer the social goals of dialog participants and presented a semi-supervised model for their classification. Their social act types were inspired by research in psychology and organizational behavior and were motivated by work in dialog understanding. They annotated a corpus in three languages using in-house annotators and achieved $\kappa$ in the range from 0.13 to 0.53.

\namecite{Georgila.et.al.2011} focused on cross-cultural aspects of persuasion or argumentation dialogs. They developed a novel annotation scheme stemming from different literature sources on negotiation and argumentation as well as from their original analysis of the phenomena. The annotation scheme is claimed to cover three dimensions of an utterance, namely speech act, topic, and response or reference to a previous utterance. They annotated 21 dialogs and reached Krippendorff's $\alpha$ between 0.38 and 0.57.

\paragraph{Summary of related work section}

Given the broad landscape of various approaches to argument analysis and persuasion studies presented in this section, we would like to stress some novel aspects of the current article. First, we aim at adapting a model of argument based on research by argumentation scholars, both theoretical and empirical. We pose several pragmatical constraints, such as register independence (generalization over several registers). Second, our emphasis is put on reliable annotations and sufficient data size (about 90k tokens). Third, we deal with fairly unrestricted Web-based sources, so additional steps of distinguishing whether the texts are argumentative are required.
Argumentation mining has been a rapidly evolving field with several major venues in 2015. We encourage readers to consult an upcoming survey article by \namecite{Lippi.Torroni.2016} or the proceedings of the 2nd Argumentation Mining workshop \cite{ARG-MINING:2015} to keep up with recent developments. However, to the best of our knowledge, the main findings of this article have not yet been made obsolete by any related work.

\section{Annotation studies and corpus creation}
\label{sec:annotation.studies}

This section describes the process of data selection, annotation, curation, and evaluation with the goal of creating a new corpus suitable for argumentation mining research in the area of computational linguistics. As argumentation mining is an evolving discipline without established and widely-accepted annotation schemes, procedures, and evaluation, we want to keep this overview detailed to ensure full reproducibility of our approach. Given the wide range of perspectives on argumentation itself \cite{vanEmeren.et.al.2014}, variety of argumentation models \cite{Bentahar.et.al.2010}, and high costs of  discourse or pragmatic annotations \cite{Prasad2008}, creating a new, reliable corpus for argumentation mining represents a substantial effort.

A motivation for creating a new corpus stems from the various use-cases discussed in the introduction, as well as some research gaps pointed in section \ref{sec:introduction} and further discussed in the survey in section \ref{sec:rel.work.arg.min} (e.g., domain restrictions, missing connection to argumentation theories, non-reported reliability or detailed schemes).

\subsection{Topics and registers}
\label{sec:topics.and.registers}

As a main field of interest in the current study, we chose controversies in education. One distinguishing feature of educational topics is their breadth of sub-topics and points of view, as they attract researchers, practitioners, parents, students, or policy-makers. We assume that this diversity leads to the linguistic variability of the education topics and thus represents a challenge for NLP. In a cooperation with researchers from the German Institute for International Educational Research\footnote{\url{http://www.dipf.de}} we identified the following current controversial topics in education in English-speaking countries: (1) \textbf{homeschooling}, (2) \textbf{public versus private schools}, (3) \textbf{redshirting} --- intentionally delaying the entry of an age-eligible child into kindergarten, allowing their child more time to mature emotionally and physically \cite{Huang.Invernizzi.2013}, (4) \textbf{prayer in schools} --- whether prayer in schools should be allowed and taken as a part of education or banned completely, (5) \textbf{single-sex education} --- single-sex classes (males and females separate) versus mixed-sex classes (``co-ed''), and (6) \textbf{mainstreaming} --- including children with special needs into regular classes.

Since we were also interested in whether argumentation differs across registers,\footnote{The distinction between registers is based on the situational context and the functional characteristics \cite[p.~6]{Biber.Conrad.2009}.} we included four different registers --- namely (1) user \textbf{comments} to newswire articles or to blog posts, (2) posts in discussion forums (\textbf{forum posts}), (3) \textbf{blog posts}, and (4) newswire \textbf{articles}.\footnote{We ignored social media sites and micro-blogs, either because searching and harvesting data is technically challenging (Facebook, Google Plus), or the texts are too short to convey argumentation, as seen in our preliminary experiments (the case of Twitter). We also did not consider debate portals (sites with \emph{pros} and \emph{cons} threads). We observed that they contain many artificial controversies or non-sense topics (for instance, createdebate.com) or their content is professionally curated (idebate.org, for example). However, we admit that debate portals might be a valuable resource in the argumentation mining research.} Throughout this work, we will refer to each article, blog post, comment, or forum posts as a \textbf{document}.
This variety of sources covers mainly user-generated content except newswire articles which are written by professionals and undergo an editing procedure by the publisher. Since many publishers also host blog-like sections on their portals, we consider as blog posts all content that is hosted on personal blogs or clearly belong to a blog category within a newswire portal.

\subsection{Raw corpus statistics}
\label{sec:raw.corpus.statistics}

\begin{figure}
\includegraphics[width=0.49\textwidth]{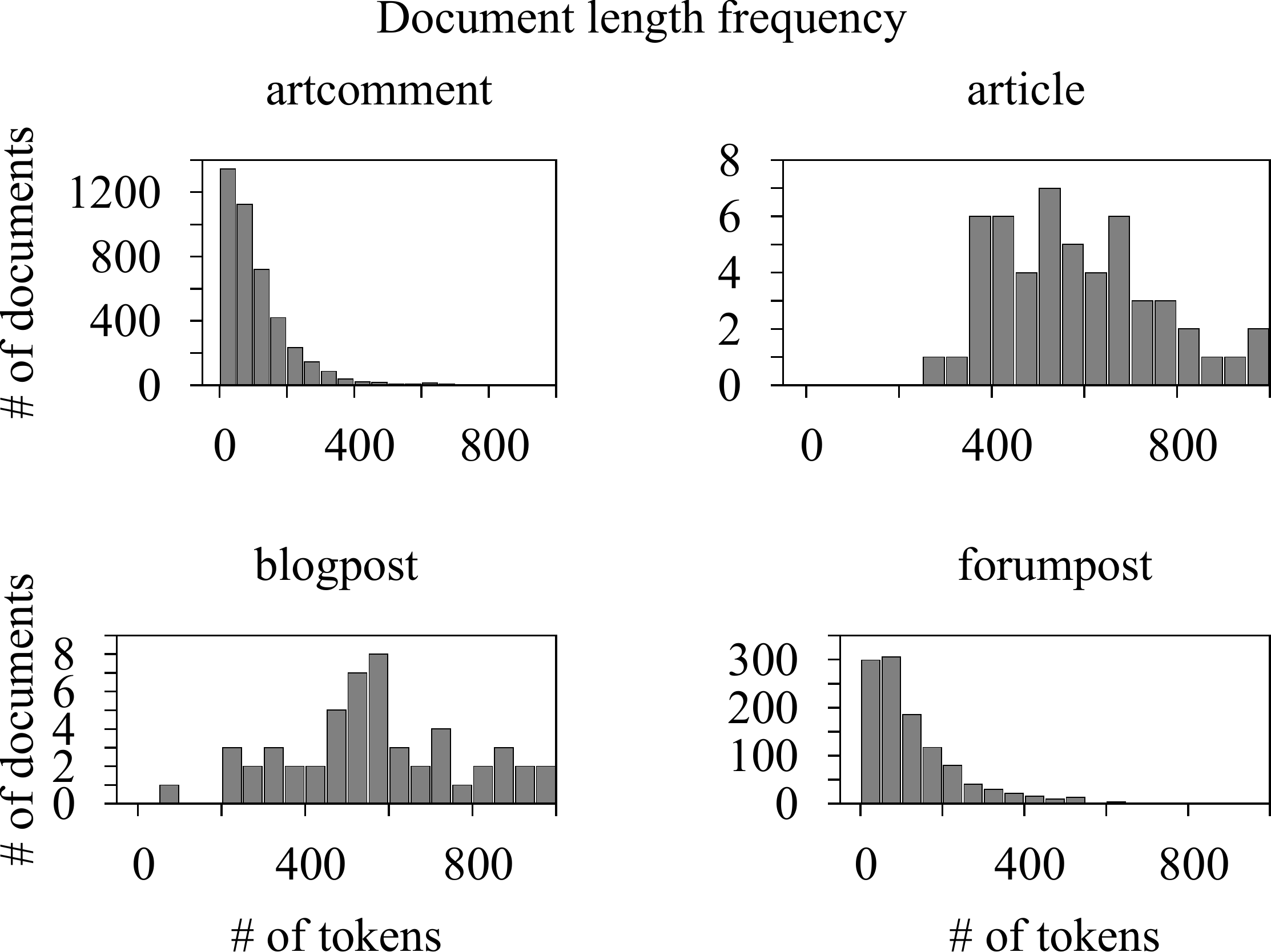}
\hfill
\includegraphics[width=0.49\textwidth]{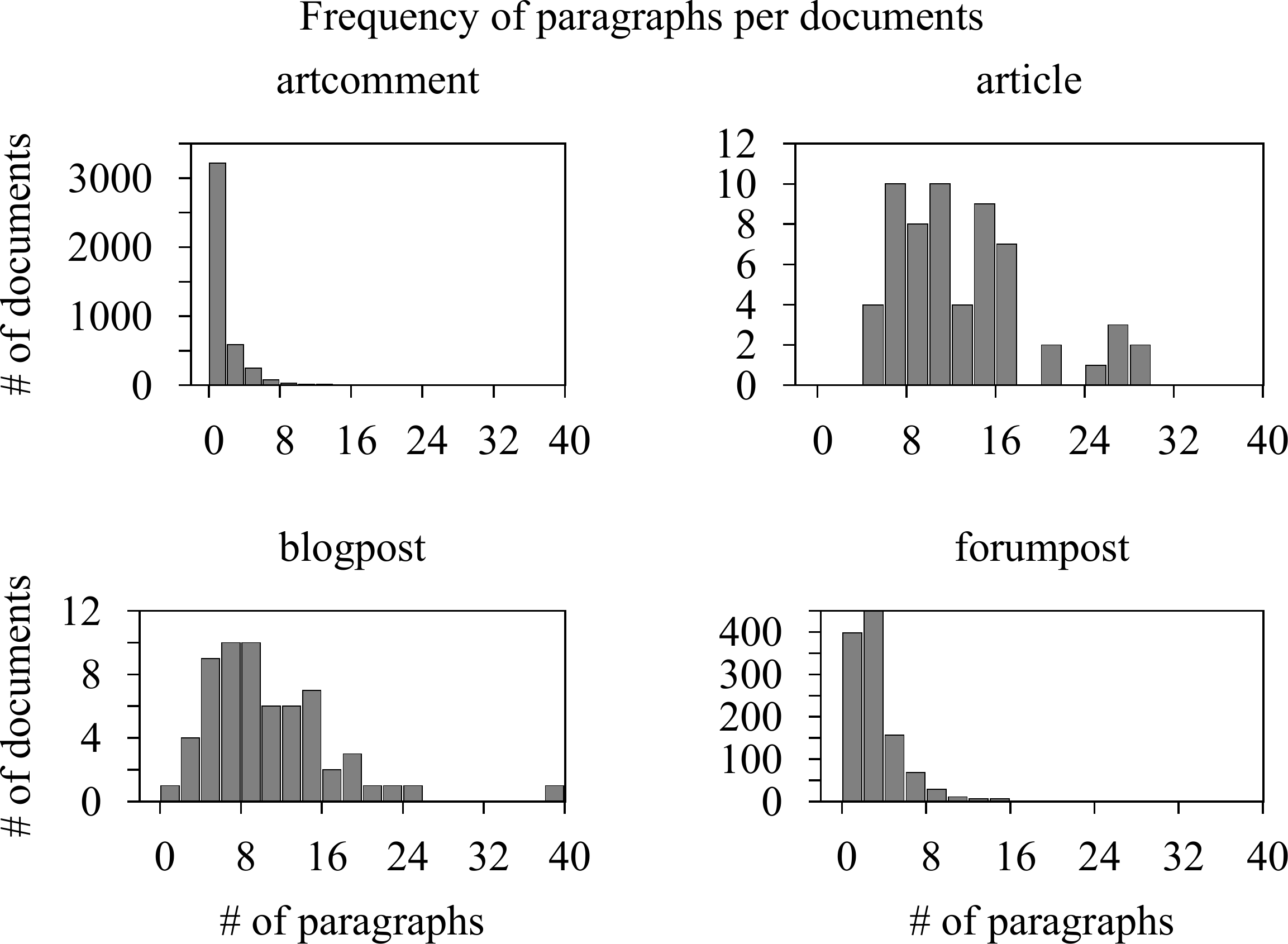}	
\caption{\label{fig:graphs-raw-corpus-lenghts} Number of documents with a certain number of tokens (left) and paragraphs (right) in the \emph{raw corpus}.}
\end{figure}

Given the six controversial topics and four different registers, we compiled a collection of plain-text documents, which we call the \emph{raw corpus}. It contains 694,110 tokens in 5,444 documents. As a coarse-grained analysis of the data, we examined the lengths and the number of paragraphs (see Figure \ref{fig:graphs-raw-corpus-lenghts}). Comments and forum posts follow a similar distribution, being shorter than 300 tokens on average. By contrast, articles and blogs are longer than 400 tokens and have 9.2 paragraphs on average. The process of compiling the \emph{raw corpus} and its further statistics are described in detail in Appendix \ref{appendix:raw.corpus.compilation}.

\subsection{Annotation study 1: Identifying persuasive documents in forums and comments}
\label{sec:annotation.study.1}

The goal of this study was to select documents suitable for a fine-grained analysis of arguments.
In a preliminary study on annotating argumentation using a small sample (50 random documents) of forum posts and comments from the \emph{raw corpus}, we found that many documents convey no argumentation at all, even in discussions about controversies. We observed that such contributions do not intend to persuade; these documents typically contain  story-sharing, personal worries, user interaction (asking questions, expressing agreement), off-topic comments, and others. Such characteristics are typical to on-line discussions in general, but they have not been examined with respect to argumentation or persuasion. Indeed, we observed that there are (1) documents that are completely unrelated and (2) documents that are related to the topic, but do not contain any argumentation. This issue has been identified among argumentation theorist; for example as \emph{external relevance} by \namecite{Paglieri.Castelfranchia.2014}.
Similar findings were also confirmed in related literature in argumentation mining, however never tackled empirically  \cite{Garcia-Villalba.Saint-Dizier.2014,Park.Cardie.2014}
These documents are thus not suitable for analyzing argumentation.

In order to filter documents that are suitable for argumentation annotation, we defined a binary document-level classification task. The distinction is made between either \emph{persuasive documents} or \emph{non-persuasive} (which includes all other sorts of texts, such as off-topic, story sharing, unrelated dialog acts, etc.).

\subsubsection{Annotation study}
\label{sec:annotation-study-1-annotation-study}

The two annotated categories were \emph{on-topic persuasive} and \emph{non-persuasive}.\footnote{
We also initially experimented with three to five categories using Likert Scale but found no extra benefits over the binary decision and thus decided to keep only two categories after the pilot experiments.
}
Three annotators with near-native English proficiency annotated a set of 990 documents (a random subset of comments and forum posts) reaching 0.59 Fleiss' $\kappa$. The final label was selected by majority voting.
The annotation study took on average of 15 hours per annotator with approximately 55 annotated documents per hour. The resulting labels were derived by majority voting. Out of 990 documents, 524 (53\%)  were labeled as on-topic persuasive. We will refer to this corpus as \emph{gold data persuasive}.

\paragraph{Sources of disagreement}
\label{sec:annotation-study1-analysis}

We examined all disagreements between annotators and discovered some typical problems, such as implicitness or topic relevance.
First, the authors often express their stance towards the topic \emph{implicitly}, so it must be inferred by the reader. To do so, certain common-ground knowledge is required. However, such knowledge heavily depends on many aspects, such as the reader's familiarity with the topic or her cultural background, as well as the context of the source website or the discussion forum thread. This also applies for sarcasm and irony.
Second, the decision whether a particular topic is \emph{persuasive} was always made with respect to the controversial topic under examination. Some authors \emph{shift the focus} to a particular aspect of the given controversy or a related issue, making the document less relevant.

\subsubsection{Discussion}
\label{sec:annotation-study1.discussion}

We achieved moderate\footnote{Following the terminology proposed by \namecite[p.~165]{Landis.Koch.1977}, although they claim that the \emph{divisions are clearly arbitrary.} For a detailed discussion on interpretation of agreement values, see for example \cite{Artstein2008}.} agreement between the annotators, although the definition of persuasiveness annotation might seem a bit fuzzy.\footnote{We also experimented with different task definition before in the preliminary studies. However, \emph{identifying argumentative documents} was misleading, as the annotators expected a reasonable argument. 
For instance, consider the following example:
 \emph{Doc\#1247 (artcomment, prayer-in-schools): ``Keep church and state separate. Period.''}
This is not an argumentative text in the traditional sense of giving reason, however, the persuasion is obvious. We are interested in all kinds of persuasive documents, not only in those that contain some clearly defined argument structures, as they can still contain useful information for decision making.
\namecite{Trabelsi.Zaiane.2014} defined a \emph{contentious document} as a document that contains expressions of one or more divergent viewpoints in response to the contention question but they did not tackle the classification of these documents. Our task also resembles aspect-based sentiment analysis (ABSA), where the aspect in our case would be the controversial topic. However, in contrast to the research in ABSA, the aspects in our case are purely abstract entities and current approaches to model ABSA do not clearly fit our task.}
We found different amounts of persuasion in the specific topics. For instance, \emph{prayer in schools} or \emph{private vs. public schools} attract persuasive discourse, while other discussed controversies often contain non-persuasive discussions, represented by \emph{redshirting} and \emph{mainstreaming}. Although these two topics are also highly controversial, the participants of on-line discussions seem to not attempt to persuade but they rather exchange information, support others in their decisions, etc. This was also confirmed by socio-psychological researchers.
\namecite{Ammari.et.al.2014} show that parents of children with special needs rely on discussion sites for accessing information and social support and that, in particular, posts containing humor, achievement, or treatment suggestions are perceived to be more socially appropriate than posts containing judgment, violence, or social comparisons. According to \namecite{Nicholson.Leask.2012}, in the online forum, parents of autistic children were seen to understand the issue because they had lived it. Assuming that participants in discussions related to young kids (e.g., redshirting, or mainstreaming) are usually females (mothers), the gender can also play a role. In a study of online persuasion, \namecite{Guadagno.Cialdini.2002} conclude that women chose to bond rather than compete (women feel more comfortable cooperating, even in a competitive environment), whereas men are motivated to compete if necessary to achieve independence.

\subsection{Annotation study 2: Annotating micro-structure of arguments}
\label{sec:annotation.study.2}

The goal of this study was to annotate documents on a detailed level with respect to an argumentation model. First, we will present the annotation scheme. Second, we will describe the annotation process. Finally, we will evaluate the agreement and draw some conclusions.

\subsubsection{Argumentation model selection}
\label{sec:argumentation.model.selection}

Given the theoretical background briefly introduced in section \ref{sec:theoretial.background}, we motivate our selection of the argumentation model by the following requirements. First, the scope of this work is to capture argumentation within a single document, thus focusing on micro-level models. Second, there should exist empirical evidence that such a model has been used for analyzing  argumentation in previous works, so it is likely to be suitable for our purposes of argumentative discourse analysis in user-generated content. 
Regarding the first requirement, two typical examples of micro-level models are the \emph{Toulmin's model} \cite{Toulmin.1958} and \emph{Walton's schemes} \cite{Walton.et.al.2008}. Let us now elaborate on the second requirement.

\paragraph{Walton's schemes}

Walton's argumentation schemes are claimed to be general and domain independent. Nevertheless, evidence from the computational linguistics field shows that the schemes lack coverage for analyzing real argumentation in natural language texts.
In examining real-world political argumentation from \cite{Walton.2005}, \namecite{Walton.2012} found out that 37.1\% of the arguments collected did not fit any of the fourteen schemes they chose so they created new schemes ad-hoc. \namecite{Cabrio2013b} selected five argumentation schemes from Walton and map these patterns to discourse relation categories in the Penn Discourse TreeBank (PDTB) \cite{Prasad2008}, but later they had to define two new argumentation schemes that they discovered in PDTB. 
Similarly, \namecite{Song.et.al.2014} admitted that the schemes are ambiguous and hard to directly apply for annotation, therefore they modified the schemes and created new ones that matched the data.

Although \namecite{Macagno.Konstantinidou.2012} show several examples of two argumentation schemes applied to few selected arguments in classroom experiments, empirical evidence presented by \namecite{Anthony.Kim.2014} reveals many practical and theoretical difficulties of annotating dialogues with schemes in classroom deliberation, providing many details on the arbitrary selection of the sub-set of the schemes, the ambiguity of the scheme definitions, concluding that the presence of the authors during the experiment was essential for inferring and identifying the argument schemes \cite[p.~93]{Anthony.Kim.2014}.

\paragraph{Toulmin's model}

Although this model (refer to section \ref{sec:toulmin.orig}) was designed to be applicable to real-life argumentation, there are numerous studies criticizing both the clarity of the model definition and the differentiation between elements of the model. \namecite{Ball1994} claims that the model can be used only for the most simple arguments and fails on the complex ones. Also \namecite{Freeman1991} and other argumentation theorists criticize the usefulness of Toulmin's framework for the description of real-life argumentative texts.
However, others have advocated the model and claimed that it can be applied to the people's ordinary argumenation \cite{Dunn.2011,Simosi2003}.

A number of studies (outside the field of computational linguistics) used Toulmin's model as their backbone argumentation framework. \namecite{Chambliss1995} experimented with analyzing 20 written documents in a classroom setting in order to find the argument patterns and parts.
\namecite{Simosi2003} examined employees' argumentation to resolve conflicts.
\namecite{Voss2006} analyzed experts' protocols dealing with problem-solving.

The model has also been used in research on computer-supported collaborative learning.
\namecite{Erduran2004} adapt Toulmin's model for coding classroom argumentative discourse among teachers and students.
\namecite{Stegmann2011} builds on a simplified Toulmin's model for scripted construction of argument in computer-supported collaborative learning.
\namecite{Garcia-Mila2013} coded utterances into categories from Toulmin's model in persuasion and  consensus-reaching among students.
\namecite{Weinberger.Fischer.2006} analyze asynchronous discussion boards in which learners engage in an argumentative discourse with the goal to acquire knowledge. For coding the argument dimension, they created a set of argumentative moves based on Toulmin's model.
Given this empirical evidence, we decided to build upon the Toulmin's model.

\subsubsection{Adaptation of Toulmin's model to argumentation in user-generated web discourse}
\label{sec:our.model}

In this annotation task, a sequence of tokens (e.g. a phrase, a sentence, or any arbitrary text span) is labeled with a corresponding argument component (such as the \emph{claim}, the \emph{grounds}, and others). There are no explicit relations between these annotation spans as the relations are implicitly encoded in the pragmatic function of the components in the Toulmin's model.

In order to prove the suitability of the Toulmin's model, we analyzed 40 random documents from the \emph{gold data persuasive} dataset using the original Toulmin's model as presented in section \ref{sec:toulmin.orig}.\footnote{The reason we are focusing on comments and forum posts on the first place is pragmatical; \namecite{Kluge2014MT} abstained from Toulmin's model when annotating long German newswire documents due to high time costs. Nevertheless, we will include another registers later in our experiments (Section \ref{sec:annotation.process}).}
We took into account sever criteria for assessment, such as frequency of occurrence of the components or their importance for the task.
We proposed some modifications of the model based on the following observations.

\paragraph{No qualifier}

Authors do not state the degree of cogency (the probability of their \emph{claim}, as proposed by Toulmin). Thus we omitted \emph{qualifier} from the model due to its absence in the data.

\paragraph{No warrant}

The \emph{warrant} as a logical explanation why one should accept the claim given the evidence is almost never stated. As pointed out by \cite[p.~92]{Toulmin.2003}, \emph{``data are appealed to explicitly, warrants implicitly.''} This observation has also been made by \namecite{Voss2006}. Also, according to \namecite[p.~205]{Eemeren.et.al.1987}, the distinction of warrant is perfectly clear only in Toulmin’s examples, but the definitions fail in practice. We omitted \emph{warrant} from the model.

\paragraph{Attacking the rebuttal}

\emph{Rebuttal} is a statement that attacks the \emph{claim}, thus playing a role of an opposing view. In reality, the authors often attack the presented rebuttals by another counter-rebuttal in order to keep the whole argument's position consistent. Thus we introduced a new component -- \emph{refutation} -- which is used for attacking the \emph{rebuttal}. Annotation of \emph{refutation} was conditioned of explicit presence of \emph{rebuttal} and enforced by the annotation guidelines. The chain \emph{rebuttal--refutation} is also known as the \emph{procatalepsis} figure in rhetoric, in which the speaker raises an objection to his own argument and then immediately answers it. By doing so, the speaker hopes to strengthen the argument by dealing with possible counter-arguments before the audience can raise them \cite[pp.~106]{Walton.2007}.

\paragraph{Implicit claim}

The claim of the argument should always reflect the main standpoint with respect to the discussed controversy. We observed that this standpoint is not always explicitly expressed, but remains implicit and must be inferred by the reader. Therefore, we allow the claim to be implicit. In such a case, the annotators must explicitly write down the (inferred) stance of the author.

\paragraph{Multiple arguments in one document}

By definition, the Toulmin's model is intended to model single argument, with the \emph{claim} in its center. However, we observed in our data, that some authors elaborate on both sides of the controversy equally and put forward an argument for each side (by \emph{argument} here we mean the \emph{claim} and its \emph{premises}, \emph{backings}, etc.). Therefore we allow multiple arguments to be annotated in one document. At the same time, we restrained the annotators from creating complex argument hierarchies.

\paragraph{Terminology} Toulmin's \emph{grounds} have an equivalent role to a \emph{premise} in the classical view on an argument \cite{vanEmeren.et.al.2014,Reed2006} in terms that they offer the reasons why one should accept the standpoint expressed by the \emph{claim}. As this terminology has been used in several related works in the argumentation mining field \cite{Stab.Gurevych.2014,Ghosh2014,Peldszus.Stede.2013,Mochales2011}, we will keep this convention and denote the \emph{grounds} as \emph{premises}.

\paragraph{The role of backing}
\label{par:backing}

One of the main critiques of the original Toulmin's model was the vague distinction between \emph{grounds}, \emph{warrant}, and \emph{backing} \cite{Freeman1991,Newman1991,Hitchcock.2003}. The role of \emph{backing} is to give additional support to the \emph{warrant}, but there is no \emph{warrant} in our model anymore. However, what we observed during the analysis, was a presence of some \emph{additional evidence}. Such evidence does not play the role of the \emph{grounds} (\emph{premises}) as it is not meant as a reason supporting the \emph{claim}, but it also does not explain the reasoning, thus is not a \emph{warrant} either. It usually supports the whole argument and is stated by the author as a certain fact. Therefore, we extended the scope of \emph{backing} as an additional support to the whole argument.

The annotators were instructed to distinguish between \emph{premises} and \emph{backing}, so that \emph{premises} should cover generally applicable reasons for the claim, whereas \emph{backing} is a single personal experience or statements that give credibility or attribute certain expertise to the author. As a sanity check, the argument should still make sense after removing \emph{backing} (would be only considered ``weaker'').

\subsubsection{Model definition}
\label{sec:toulmin.model.definition}

We call the model as a \emph{modified Toulmin's model}. It contains five argument components, namely \emph{claim}, \emph{premise}, \emph{backing}, \emph{rebuttal}, and \emph{refutation}. When annotating a document, any arbitrary token span can be labeled with an argument component; the components do not overlap. The spans are not known in advance and the annotator thus chooses the span and the component type at the same time. All components are optional (they do not have to be present in the argument) except the \emph{claim}, which is either explicit or implicit (see above). If a token span is not labeled by any argument component, it is not considered as a part of the argument and is later denoted as \emph{none} (this category is not assigned by the annotators).

An example analysis of a forum post is shown in Figure \ref{example:arg.annotation1}. Figure \ref{fig:scheme.example.toulmin.extended} then shows a diagram of the analysis from that example (the content of the argument components was shortened or rephrased).

\begin{figure}
\begin{quote}
\textbf{Doc\#4733 (forumpost, public-private-schools)} 
\claim{
The public schooling system is not as bad as some may think.} \rebuttal{
Some mentioned that those who are educated in the public schools are less educated,} \refutation{
well I actually think it would be in the reverse.} \premise{
Student who study in the private sector actually pay a fair amount of fees to do so and I believe that the students actually get let off for a lot more than anyone would in a public school. And its all because of the money.\P\newline
In a private school, a student being expelled or suspended is not just one student out the door, its the rest of that students schooling life fees gone. Whereas in a public school, its just the student gone.}\P\newline
\backing{
I have always gone to public schools and when I finished I got into University. I do not feel disadvantaged at all.}
\end{quote}
\caption{\label{example:arg.annotation1} An annotation example using the \emph{modified Toulmin's model}.}
\end{figure}

\begin{figure}
\centering
\includegraphics[width=0.7\textwidth]{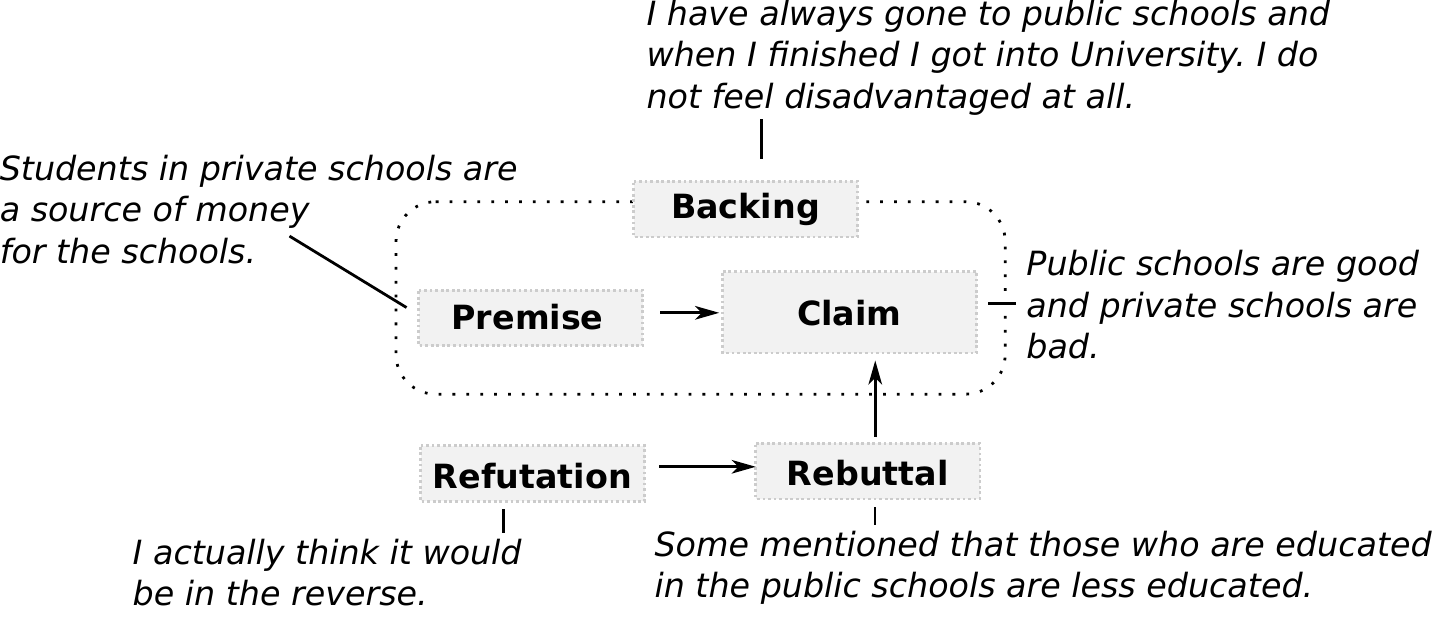}
\caption{\label{fig:scheme.example.toulmin.extended} Modified Toulmin's model used for annotation of arguments with an instantiated example from a single discussion forum post on \emph{public vs. private schools} (see Figure \ref{example:arg.annotation1}). The arrows show relations between argument components; the relations are implicit and inherent in the model. By contrast to the example of \emph{original Toulmin's model} in Figure \ref{fig:toulmin-orig}, we do not propose any connective phrases to the relations (such as \emph{so}, \emph{unless}, etc.).}
\end{figure}

\subsubsection{Annotation workflow}
\label{sec:annotation.process}

The annotation experiment was split into three phases. All documents were annotated by three independent annotators, who participated in two training sessions.
During the \textbf{first phase}, 50 random comments and forum posts were annotated. Problematic cases were resolved after discussion and the guidelines were refined. In the \textbf{second phase}, we wanted to extend the range of annotated registers, so we selected 148 comments and forum posts as well as 41 blog posts.
After the second phase, the annotation guidelines were final.\footnote{The annotation guidelines are available under CC-BY-SA license at \url{https://www.ukp.tu-darmstadt.de/data/argumentation-mining/}}

In the \textbf{final phase}, we extended the range of annotated registers and added newswire articles from the \emph{raw corpus} in order to test whether the annotation guidelines (and inherently the model) is general enough. Therefore we selected 96 comments/forum posts, 8 blog posts, and 8 articles for this phase. A detailed inter-annotator agreement study on documents from this final phase will be reported in section \ref{sec:toulmin.IAA}.

The annotations were very time-consuming. In total, each annotator spent 35 hours by annotating in the course of five weeks. Discussions and consolidation of the gold data took another 6 hours. Comments and forum posts required on average of 4 minutes per document to annotate, while blog posts and articles on average of 14 minutes per document.
Examples of annotated documents from the gold data are listed in Appendix \ref{app:toulmin.examples}.

\paragraph{Discarding documents during annotation}

We discarded 11 documents out of the total 351 annotated documents. Five forum posts, although annotated as \emph{persuasive} in the first annotation study, were at a deeper look a mixture of two or more posts with missing quotations,\footnote{All of them came from the same source website that does not support any HTML formatting of quotations.} therefore unsuitable for analyzing argumentation. Three blog posts and two articles were found not to be argumentative (the authors took no stance to the discussed controversy) and one article was an interview, which the current model cannot capture (a dialogical argumentation model would be required).

For each of the 340 documents, the gold standard annotations were obtained using the majority vote. If simple majority voting was not possible (different boundaries of the argument component together with a different component label), the gold standard was set after discussion among the annotators. We will refer to this corpus as the \emph{gold standard Toulmin} corpus. The distribution of topics and registers in this corpus in shown in Table \ref{tab:toulmin.gold.topic.register.distribution}, and Table \ref{tab:stats-corpus-phase2-1} presents some lexical statistics.

\begin{table}
\begin{tabular}{l|rrrr|l}
\textbf{Topic} $\backslash$ \textbf{Register} & \textbf{Comment} & \textbf{Forum post} & \textbf{Blog post} & \textbf{Article} & \textbf{Total} \\ \hline
Homeschooling	&32	&12	&11	&1	&56\\
Mainstreaming	&12	&5	&3	&1	&21\\
Prayer in schools	&31	&14	&10	&0	&55\\
Public vs. private	&117	&10	&7	&0	&134\\
Redshirting	&19	&13	&4	&1	&37\\
Single-sex education	&14	&12	&9	&2	&37\\ \hline
Total & 216 & 73 & 46 & 5 & 340
\end{tabular}
\caption{\label{tab:toulmin.gold.topic.register.distribution} Topic and register distribution in the \emph{gold standard Toulmin} corpus.}
\end{table}

\begin{table}
\begin{tabular}{l|rr|rr}
\textbf{Register} & \textbf{Tokens} & \textbf{Mean} & \textbf{Sentences} & \textbf{Mean} \\ \hline
Comments & 35,461 & 164.17 $\pm$ 155.87 & 1,748 & 8.09 $\pm$ 7.68 \\
Forums posts & 13,033 & 178.53 $\pm$ 132.33 & 641 & 8.78 $\pm$ 7.53 \\
Blogs & 32,731 & 711.54 $\pm$ 293.72 & 1,378 & 29.96 $\pm$ 14.82 \\
Articles & 3,448 & 689.60 $\pm$ 183.34 & 132 & 24.60 $\pm$ 6.58 \\ \hline
All & 84,673 & 249.04 $\pm$ 261.77 & 3,899 & 11.44 $\pm$ 11.70
\end{tabular}
\caption{\label{tab:stats-corpus-phase2-1} \emph{Gold standard Toulmin} corpus statistics.}
\end{table}

\subsubsection{Annotation set-up}
\label{sec:annotation-set-up}

Based on pre-studies, we set the minimal unit for annotation as \emph{token}.\footnote{We also considered sentences or clauses. The sentence level seems to be reasonable in most of the cases, however, it is too coarse-grained if a sentence contains multiple clauses that belong to different argumentation components.
Segmentation to clauses is not trivial and has been considered as a separate task since CoNLL 2001 \cite{Tjong.et.al.2001}. Best systems based on Join-CRF reach 0.81 $F_1$ score \cite{Nguyen.et.al.2009} for embedded clauses and 0.92 for non-embedded \cite{Zhang.et.al.2013}. To the best of our knowledge, there is no available out-of-box solution for clause segmentation, thus we took \emph{sentences} as another level of segmentation. Nevertheless, pre-segmenting the text to clauses and their relation to argument components deserves future investigation.}
The documents were pre-segmented using the Stanford Core NLP sentence splitter \cite{Manning.et.al.2014} embedded in the DKPro Core framework \cite{Eckart.Gurevych.2014}. Annotators were asked to stick to the sentence level by default and label entire pre-segmented sentences. They should switch to annotations on the token level only if (a) a particular sentence contained more than one argument component, or (b) if the automatic sentence segmentation was wrong. Given the ``noise'' in user-generated Web data (wrong or missing punctuation, casing, etc.), this was often the case.

Annotators were also asked to rephrase (summarize) each annotated argument component into a simple statement when applicable, as shown in Figure \ref{fig:scheme.example.toulmin.extended}. This was used as a first sanity checking step, as each argument component is expected to be a coherent discourse unit. For example, if a particular occurrence of a \emph{premise} cannot be summarized/rephrased into one statement, this may require further splitting into two or more \emph{premises}.

For the actual annotations, we developed a custom-made web-based application that allowed users to switch between different granularity of argument components (tokens or sentences), to annotate the same document in different argument ``dimensions'' (logos and pathos), and to write summary for each annotated argument component.

\subsubsection{Inter-annotator agreement}
\label{sec:toulmin.IAA}

As a measure of annotation reliability, we rely on Krippendorff's unitized alpha ($\alphaunit$) \cite{Krippendorff.2004}. To the best of our knowledge, this is the only agreement measure that is applicable when both \emph{labels} and \emph{boundaries} of segments are to be annotated.

Although the measure has been used in related annotation works \cite{Ghosh2014,Stab.Gurevych.2014,Kluge2014MT}, there is one important detail that has not been properly communicated. The $\alphaunit$ is computed over a \emph{continuum} of the smallest units, such as tokens. This continuum corresponds to a single document in the original Krippendorff's work. However, there are two possible extensions to multiple documents (a corpus), namely (a) to compute $\alphaunit$ for each document first and then report an average value, or (b) to concatenate all documents into one large continuum and compute $\alphaunit$ over it.
The first approach with averaging yielded extremely high the standard deviation of $\alphaunit$  (i.e., avg. = 0.253; std. dev. = 0.886; median = 0.476 for the \emph{claim}). This says that some documents are easy to annotate while others are harder, but interpretation of such averaged value has no evidence either in \cite{Krippendorff.2004} or other papers based upon it. Thus we use the other methodology and treat the whole corpus as a single long continuum (which yields in the example of \emph{claim} 0.541 $\alphaunit$).\footnote{Another pitfall of the $\alphaunit$ measure when documents are concatenated to create a single continuum is that its value depends on the order of the documents (the annotated spans, respectively). We did the following experiment: using 10 random annotated documents, we created all 362,880 possible concatenations and measured the $\alphaunit$ for each permutation. The resulting standard error was 0.002, so the influence of the ordering is rather low. Still, each reported $\alphaunit$ was averaged from 100 random concatenations of the analyzed documents.}

\begin{table}
\begin{tabular}{ll|llllll}
  & \textbf{All topics}	&\textbf{HS}	&\textbf{RS}	&\textbf{PIS}	&\textbf{SSE}	&\textbf{MS}	&\textbf{PPS}\\ \hline
\multicolumn{8}{c}{  } \\ 
\multicolumn{8}{l}{(a) Comments + Forum posts} \\ \hline
Claim	&\textbf{0.59}	&\textbf{0.52}	&0.36	&\textbf{0.70}	&\textbf{0.69}	&\textbf{0.51}	&\textbf{0.55}\\
Premise	&\textbf{0.69}	&0.35	&0.31	&\textbf{0.80}	&0.47	&0.16	&0.38\\
Backing	&0.48	&0.15	&0.06	&0.36	&\textbf{0.54}	&0.14	&0.49\\
Rebuttal	&0.37	&0.12	&0.03	&0.25	&-0.02	&\textbf{0.80}	&0.34\\
Refutation	&0.08	&0.03	&-0.01	&-0.02	&--	&0.32	&0.11\\
\textbf{Joint logos}	&\textbf{0.60}	&0.28	&0.19	&\textbf{0.68}	&0.49	&0.16	&0.44\\ \hline
\multicolumn{8}{c}{  } \\ 
\multicolumn{8}{l}{(b) Articles + Blog posts} \\ \hline
Claim	&0.22	&-0.02	&-0.03	&--	&--	&0.33	&--\\
Premise	&0.24	&0.02	&0.24	&--	&-0.04	&0.40	&--\\
Backing	&-0.03	&0.18	&-0.20	&--	&0.26	&-0.20	&--\\
Rebuttal	&0.01	&--	&0.08	&--	&-0.08	&-0.08	&--\\
Refutation	&0.34	&--	&0.40	&--	&-0.01	&-0.01	&--\\
\textbf{Joint logos}	&0.09	&0.05	&0.01	&--	&0.08	&0.04	&--\\ \hline
\multicolumn{8}{c}{  } \\ 
\multicolumn{8}{l}{(c) Articles + Blog posts + Comments + Forum posts} \\ \hline
Claim	&\textbf{0.54}	&\textbf{0.52}	&0.28	&\textbf{0.70}	&\textbf{0.71}	&0.43	&\textbf{0.55}\\
Premise	&\textbf{0.62}	&0.33	&0.29	&\textbf{0.80}	&0.32	&0.27	&0.38\\
Backing	&0.31	&0.17	&0.00	&0.36	&0.42	&-0.04	&0.49\\
Rebuttal	&0.08	&0.12	&0.09	&0.25	&0.02	&0.03	&0.34\\
Refutation	&0.17	&0.03	&0.38	&-0.02	&0.00	&0.20	&0.11\\ \hline
\textbf{Joint logos}	& 0.48	&0.27	&0.14	&\textbf{0.68}	&0.34	&0.08	&0.44\\

\end{tabular}
\caption{\label{tab:iaa.study2} Inter-annotator agreement (Krippendorff's $\alphaunit$) across various registers, topics, and argument components. Bold values emphasize $\alphaunit \geq 0.50$. \textbf{Joint logos} is a joint $\alphaunit$ for all argument components in the logos dimension (\emph{claim}, \emph{premise}, \emph{backing}, \emph{rebuttal}, \emph{refutation}). \textbf{HS} -- homeschooling, \textbf{RS} -- redshirting, \textbf{PIS} -- prayer in schools, \textbf{SSE} -- single sex education, \textbf{MS} -- mainstreaming, \textbf{PPS} -- private vs. public schools.}
\end{table}

Table \ref{tab:iaa.study2} shows the inter-annotator agreement as measured on documents from the last annotation phase (see section \ref{sec:annotation.process}). The overall $\alphaunit$ for all register types, topics, and argument components is 0.48 in the \emph{logos} dimension (annotated with the \emph{modified Toulmin's model}). Such agreement can be considered as moderate by the measures proposed by \namecite{Landis.Koch.1977}, however, direct interpretation of the agreement value lacks consensus \cite[p.~591]{Artstein2008}. Similar inter-annotator agreement numbers were achieved in the relevant works in argumentation mining (refer to Table \ref{tab:related-work} in section \ref{sec:rel.work.arg.min}; although most of the numbers are not directly comparable, as different inter-annotator metrics were used on different tasks).

There is a huge difference in $\alphaunit$ regarding the registers between comments + forums posts ($\alphaunit$ 0.60, Table \ref{tab:iaa.study2}a) and articles + blog posts ($\alphaunit$ 0.09, Table \ref{tab:iaa.study2}b) in the logos dimension. If we break down the value with respect to the individual argument components, the agreement on \emph{claim} and \emph{premise} is substantial in the case of comments and forum posts (0.59 and 0.69, respectively). By contrast, these argument components were annotated only with a fair agreement in articles and blog posts (0.22 and 0.24, respectively).

As can be also observed from Table \ref{tab:iaa.study2}, the annotation agreement in the logos dimension varies regarding the document topic. While it is substantial/moderate for \emph{prayer in schools} (0.68) or \emph{private vs. public schools} (0.44), for some topics it remains rather slight, such as in the case of \emph{redshirting} (0.14) or \emph{mainstreaming} (0.08).

\subsubsection{Causes of disagreement -- quantitative analysis}

First, we examine the disagreement in annotations by posing the following research question: \emph{are there any measurable properties of the annotated documents that might systematically cause low inter-annotator agreement?} We use Pearson's correlation coefficient between $\alphaunit$ on each document and the particular property under investigation.
We investigated the following set of measures.

\begin{itemize}
\item \emph{Full sentence coverage ratio} represents a ratio of argument component boundaries that are aligned to sentence boundaries. The value is 1.0 if all annotations in the particular document are aligned to sentences and 0.0 if no annotations match the sentence boundaries. Our hypothesis was that automatic segmentation to sentences was often incorrect, therefore annotators had to switch to the token level annotations and this might have increased disagreement on boundaries of the argument components.
\item \emph{Document length}, \emph{paragraph length} and \emph{average sentence length}. Our hypotheses was that the length of documents, paragraphs, or sentences negatively affects the agreement.
\item \emph{Readability measures}. We tested four standard readability measures, namely \emph{Ari} \cite{Senter.Smith.1967}, \emph{Coleman-Liau} \cite{Coleman.Liau.1975}, \emph{Flesch} \cite{Flesch.1948}, and \emph{Lix} \cite{Bjornsonn.1968} to find out whether readability of the documents plays any role in annotation agreement.
\end{itemize}

\begin{table}
\begin{tabular}{lrrrrrrrr}
& \textbf{SC} & \textbf{DL} & \textbf{APL} & \textbf{ASL} & \textbf{ARI} & \textbf{C-L} & \textbf{Flesch} & \textbf{LIX} \\ \hline
all data & -0.14 & -0.14 & 0.01 & 0.04 & 0.07 & 0.08 & -0.11 & 0.07  \\ \hline
comments & -0.17 & \textbf{-0.64} & 0.13 & 0.01 & 0.01 & 0.01 & -0.11 & 0.01  \\
forum posts & -0.08 & -0.03 & -0.08 & -0.03 & 0.08 & 0.24 & -0.17 & 0.20 \\
blog posts & -0.50 & 0.21 & \textbf{-0.81} & -0.61 & -0.39 & 0.47 & 0.04 & -0.07 \\
articles & 0.00 & -0.64 & -0.43 & -0.65 & -0.25 & 0.39 & -0.27 & -0.07  \\ \hline
homeschooling & -0.10 & -0.29 & -0.18 & 0.34 & 0.35 & 0.31 & -0.38 & 0.46  \\
redshirting & -0.16 & 0.07 & -0.26 & -0.07 & 0.02 & 0.14 & -0.06 & -0.09 \\
prayer-in-school & -0.24 & \textbf{-0.85} & 0.30 & 0.07 & 0.14 & 0.11 & -0.25 & 0.24  \\
single sex & -0.08 & -0.36 & -0.28 & -0.16 & -0.17 & 0.05 & 0.06 & 0.06  \\
mainstreaming & -0.27 & -0.00 & -0.03 & 0.06 & 0.20 & 0.29 & -0.19 & 0.03  \\
public private & 0.18 & 0.19 & 0.30 & -0.26 & \textbf{-0.58} & \textbf{-0.51} & \textbf{0.51} & \textbf{-0.56} \\
\end{tabular}
\caption{\label{tab:correlations-gold} Correlations between $\alphaunit$ and various measures on different data sub-sets. \textbf{SC} -- full sentence coverage; \textbf{DL} -- document length; \textbf{APL} -- average paragraph length; \textbf{ASL} = average sentence length; \textbf{ARI}, \textbf{C-L (Coleman-Liau)}, \textbf{Flesch}, \textbf{LIX} -- readability measures. Bold numbers denote statistically significant correlation ($p < 0.05$).}
\end{table}

Correlation results are listed in Table \ref{tab:correlations-gold}. We observed the following statistically significant ($p < 0.05$) correlations. First, \emph{document length} negatively correlates with agreement in \emph{comments}. The longer the comment was the lower the agreement was. Second, \emph{average paragraph length} negatively correlates with agreement in \emph{blog posts}. The longer the paragraphs in blogs were, the lower agreement was reached. Third, all \emph{readability scores} negatively correlate with agreement in the \emph{public vs. private school} domain, meaning that the more complicated the text in terms of readability is, the lower agreement was reached. We observed no significant correlation in \emph{sentence coverage} and \emph{average sentence length} measures. We cannot draw any general conclusion from these results, but we can state that some registers and topics, given their properties, are more challenging to annotate than others.

\paragraph{Probabilistic confusion matrix}

Another qualitative analysis of disagreements between annotators was performed by constructing a \emph{probabilistic confusion matrix} \cite{Cinkova.et.al.2012} on the token level.\footnote{\emph{``Properties: The sum in any row is 1. The $j$-th row of the matrix contains probabilities of assigning $t_i$ given that another annotator has chosen $t_j$ for the same instance. Thus, the $j$-th row of matrix describes expected tagging confusion related to the tag $t_j$.''} \cite[p.~846]{Cinkova.et.al.2012}} The biggest disagreements, as can be seen in Table \ref{tab:probabilistic-confusion-matrix}, is caused by \emph{rebuttal} and \emph{refutation} confused with \emph{none} (0.27 and 0.40, respectively). This is another sign that these two argument components were very hard to annotate. As shown in Table \ref{tab:iaa.study2}, the $\alphaunit$ was also low -- 0.08 for \emph{rebuttal} and 0.17 for \emph{refutation}.

\begin{table}
\begin{tabular}{l|rrrrrr}
	& \textbf{Claim}	& \textbf{Premise}	& \textbf{Backing}	& \textbf{Rebuttal}	& \textbf{Refutation}	& \textbf{None}\\ \hline
\textbf{Claim}	& 0.59	& 0.17	& 0.07	& 0.04	& 0.02	& 0.11\\
\textbf{Premise}	& 0.01	& 0.54	& 0.16	& 0.06	& 0.00	& 0.23\\
\textbf{Backing}	& 0.03	& 0.17	& 0.52	& 0.02	& 0.00	& 0.25\\
\textbf{Rebuttal}	& 0.16	& 0.12	& 0.10	& 0.32	& 0.03	& 0.27\\
\textbf{Refutation}	& 0.05	& 0.19	& 0.00	& 0.23	& 0.13	& 0.40\\
\textbf{None}	& 0.01	& 0.12	& 0.16	& 0.02	& 0.00	& 0.69\\
\end{tabular}
\caption{\label{tab:probabilistic-confusion-matrix} Probabilistic confusion matrix between all annotators.}
\end{table}

\subsubsection{Causes of disagreement -- qualitative analysis and problematic phenomena}
\label{sec:iaa.toulmin.causes.of.disagreement.qualitative}

We analyzed the annotations and found the following phenomena that usually caused disagreements between annotators.

\paragraph{Granularity of argument components}

Each argument component (e.g., \emph{premise} or \emph{backing}) should express one consistent and coherent piece of information, for example a single reason in case of the \emph{premise} (see Section \ref{sec:annotation-set-up}).
However, the decision whether a longer text should be kept as a single argument component or segmented into multiple components is subjective and highly text-specific.\footnote{An example from Doc\#4566 (artcomment, public-private-schools): One annotator labeled two premises: \premise{I send my kids to public schools because I care about them - their links with their diverse local community etc - but also because I care about the kind of culture they live in.} (summary: In public schools, kids have links with community and culture) \premise{To me, learning to care about and contribute to society as a whole - not just your own personal interests - is the best value a child can inherit.} (summary: At public, kids learn to contribute as a society). Another annotator labeled the same text as one single premise: \premise{I send my kids to public schools because I care about them - their links with their diverse local community etc - but also because I care about the kind of culture they live in. To me, learning to care about and contribute to society as a whole - not just your own personal interests - is the best value a child can inherit.} (summary: Kids should learn the cultural diversity).}

\paragraph{Rhetorical questions}

While rhetorical questions have been researched extensively in linguistics \cite{Schmidt.1977,Han2002,Egg2007,Lee-Goldman2006}, their role in argumentation represents a substantial research question \cite{Roberts.Kreuz.1994,Ottati1.et.al.1999,Petty.et.al.1981,Frank.1990,Ilie1999}. \namecite{Teninbaum.2011} provides a brief history of rhetorical questions in persuasion. In short, rhetorical questions should provoke the reader.
From the perspective of our argumentation model, rhetorical questions might fall both into the \emph{logos} dimension (and thus be labeled as, e.g., \emph{claim}, \emph{premise}, etc.) or into the \emph{pathos} dimension (refer to Section \ref{sec:dimensions.of.argument}). Again, the decision is usually not clear-cut.

\paragraph{Refutation versus premise}

As introduced in section \ref{sec:our.model}, \emph{rebuttal} attacks the \emph{claim} by presenting an opponent's view. In most cases, the \emph{rebuttal} is again attacked by the author using \emph{refutation}. From the pragmatic perspective, \emph{refutation} thus supports the author's stance expressed by the \emph{claim}. Therefore, it can be easily confused with \emph{premises}, as the function of both is to provide support for the \emph{claim}. \emph{Refutation} thus only takes place if it is meant as a reaction to the \emph{rebuttal}. It follows the discussed matter and contradicts it. Such a discourse is usually expressed as:

\begin{quote}
\claim{My claim.} \rebuttal{On the other hand, some people claim XXX which makes my claim wrong.} \refutation{But this is not true, because of YYY.}
\end{quote}

However, the author might also take the following defensible approach to formulate the argument:

\begin{quote}
\rebuttal{Some people claim XXX-1 which makes my claim wrong.} \refutation{But this is not true, because of YYY-1.}
\rebuttal{Some people claim XXX-2 which makes my claim wrong.} \refutation{But this is not true, because of YYY-2.}
\claim{Therefore my claim.}
\end{quote}

If this argument is formulated without stating the \emph{rebuttals}, it would be equivalent to the following:

\begin{quote}
\premise{YYY-1.} \premise{YYY-2.} \claim{Therefore my claim.}
\end{quote}

This example shows that \emph{rebuttal} and \emph{refutation} represent a rhetorical device to produce arguments, but the distinction between \emph{refutation} and \emph{premise} is context-dependent and on the functional level both \emph{premise} and \emph{refutation} have very similar role -- to support the author's standpoint.
Although introducing dialogical moves into monological model and its practical consequences, as described above, can be seen as a shortcoming of our model, this rhetoric figure has been identified by argumentation researchers as \emph{procatalepsis} \cite[pp.~106]{Walton.2007}. 
A broader view on incorporating opposing views (or lack thereof) is discussed under the term \emph{confirmation bias} by \cite[p.~63]{Mercier.Sperber.2011} who claim that \emph{``[...] people are trying to convince others. They are typically looking for arguments and evidence to confirm their own claim, and ignoring negative arguments and evidence unless they anticipate having to rebut them.''}
The dialectical attack of possible counter-arguments may thus strengthen one's own argument.

One possible solution would be to refrain from capturing this phenomena completely and to simplify the model to claims and premises, for instance. However, the following example would then miss an important piece of information, as the last two clauses would be left un-annotated. At the same time, annotating the last clause as \emph{premise} would be misleading, because it does not support the \emph{claim} (in fact, it supports it only indirectly by attacking the \emph{rebuttal}; this can be seen as a support is considered as an admissible extension of abstract argument graph by \cite{Dung.1995}).

\begin{quote}
\textbf{Doc\#422 (forumpost, homeschooling)}
\claim{I try not to be anti-homeschooling, but... it's just hard for me.} \premise{I really haven't met any homeschoolers who turned out quite right, including myself.} I apologize if what I'm saying offends any of you - that's not my intention, \rebuttal{I know that there are many homeschooled children who do just fine,} but \refutation{that hasn't been my experience.}
\end{quote}

To the best of our knowledge, these context-dependent dialogical properties of argument components using Toulmin's model have not been solved in the literature on argumentation theory and we suggest that these observations should be taken into account in the future research in monological argumentation.

\paragraph{Purely sarcastic argumentation and fallacies in general}

Appeal to emotion, sarcasm, irony, or jokes are common in argumentation in user-generated Web content. We also observed documents in our data that were purely sarcastic (the \emph{pathos} dimension), therefore logical analysis of the argument (the \emph{logos} dimension) would make no sense. However, given the structure of such documents, some \emph{claims} or \emph{premises} might be also identified. Such an argument is a typical example of fallacious argumentation, which intentionally \emph{pretends} to present a valid argument, but its persuasion is conveyed purely for example by appealing to emotions of the reader \cite{Tindale.2007}.

\subsubsection{Analysis of annotated corpus from argumentation research perspective}
\label{sec:annotation.study.1.analysis.of.corpus}

We present some statistics of the annotated data that are important from the argumentation research perspective. Regardless of the register, 48\% of \emph{claims} are implicit. This means that the authors assume that their standpoint towards the discussed controversy can be inferred by the reader and give only reasons for that standpoint. Also, explicit \emph{claims} are mainly written just once, only in 3\% of the documents the \emph{claim} was rephrased and occurred multiple times.

In 6\% of the documents, the reasons for an implicit \emph{claim} are given only in the \emph{pathos} dimension, making the argument purely persuasive without logical argumentation.

The ``myside bias'', defined as a bias against information supporting another side of an argument \cite{Perkins.1985,Wolfe.et.al.2009}, can be observed by the presence of \emph{rebuttals} to the author's \emph{claim} or by formulating arguments for both sides when the overall stance is neutral. While 85\% of the documents do not consider any opposing side, only 8\% documents present a \emph{rebuttal}, which is then attacked by \emph{refutation} in 4\% of the documents. Multiple \emph{rebuttals} and \emph{refutations} were found in 3\% of the documents. Only 4\% of the documents were overall neutral and presented arguments for both sides, mainly in blog posts.

\paragraph{Hedging in claims}

We were also interested whether mitigating linguistic devices are employed in the annotated arguments, namely in their main stance-taking components, the \emph{claims}. Such devices typically include parenthetical verbs, syntactic constructions, token agreements, hedges, challenge questions, discourse markers, and tag questions, among others \cite{Flores-Ferran.Lovejoy.2015}. In particular, \namecite[p.~1]{Kaltenbock.et.al.2010} define hedging as \emph{a discourse strategy that reduces the force or truth of an utterance and thus reduces the risk a speaker runs when uttering a strong or firm assertion or other speech act.} We manually examined the use of hedging in the annotated \emph{claims}.

Our main observation is that hedging is used differently across topics. For instance, about 30-35\% of claims in \emph{homeschooling} and \emph{mainstreaming} signal the lack of a full commitment to the expressed stance, in contrast to \emph{prayer in schools} (15\%) or \emph{public vs. private schools} (about 10\%). Typical hedging cues include speculations and modality (\emph{``\textbf{If} I have kids, I will \textbf{probably} homeschool them.''}), statements as neutral observations (\emph{``\textbf{It's not wrong to hold the opinion that in general it's better} for kids to go to school than to be homeschooled.''}), or weasel\footnote{\url{https://en.wikipedia.org/wiki/Weasel_word}} phrases \cite{Farkas.et.al.2010} (\emph{``\textbf{In some cases}, inclusion can work fantastically well.'', ``\textbf{For the majority of the children in the school}, mainstream would not have been a suitable placement.''}).

On the other hand, most \emph{claims} that are used for instance in the \emph{prayer in schools} arguments are very direct, without trying to diminish its commitment to the conveyed belief (for example, \emph{``NO PRAYER IN SCHOOLS!... period.''}, \emph{``Get it out of public schools''}, \emph{``Pray at home.''}, or \emph{``No organized prayers or services anywhere on public school board property - FOR ANYONE.''}). Moreover, some claims are clearly offensive, persuading by direct imperative clauses towards the opponents/audience (\emph{``TAKE YOUR KIDS PRIVATE IF YOU CARE AS I DID''}, \emph{``Run, don't walk, to the nearest private school.''}) or even accuse the opponents for taking a certain stance (\emph{``You are a bad person if you send your children to private school.''}).

These observations are consistent with the findings from the first annotation study on persuasion (see section \ref{sec:annotation-study1.discussion}), namely that some topics attract \emph{heated} argumentation where participant take very clear and reserved standpoints (such as \emph{prayer in schools} or \emph{private vs. public schools}), while discussions about other topics are rather milder.
It has been shown that the choices a speaker makes to express a position are informed by their social and cultural background, as well as their ability to speak the language \cite{Kreutel.2007,Dippold.2007,Flores-Ferran.Lovejoy.2015}. However, given the uncontrolled settings of the user-generated Web content, we cannot infer any similar conclusions in this respect.

\paragraph{Analyzing type of support}

We investigated \emph{premises} across all topics in order to find the type of support used in the argument. We followed the approach of \namecite{Park.Cardie.2014}, who distinguished three types of propositions in their study, namely \emph{unverifiable}, \emph{verifiable non-experiential}, and \emph{verifiable experiential}.

\emph{Verifiable non-experiential} and \emph{verifiable experiential} propositions, unlike \emph{unverifiable propositions}, contain an objective assertion, where objective means ``expressing or dealing with facts or conditions as perceived without distortion by personal feelings, prejudices, or interpretations.''\footnote{\url{http://www.merriam-webster.com/dictionary/objective}}  Such assertions have truth values that can be proved or disproved with objective evidence; the correctness of the assertion or the availability of the objective evidence does not matter \cite[p.~31]{Park.Cardie.2014}.
A verifiable proposition can further be distinguished as experiential or not, depending on whether the proposition is about the writer's personal state or experience or something non-experiential. Verifiable experiential propositions are sometimes referred to as anectotal evidence, provide the novel knowledge that readers are seeking \cite[p.~31]{Park.Cardie.2014}.

Table \ref{tab:support} shows the distribution of the premise types with examples for each topic from the annotated corpus. As can be seen in the first row, arguments in \emph{prayer in schools} contain majority (73\%) of unverifiable premises. Closer examination reveals that their content vary from general vague propositions to obvious fallacies, such as a hasty generalization, straw men, or slippery slope. As \namecite{Nieminen.Mustonen.2014} found out, fallacies are very common in argumentation about religion-related issues.
On the other side of the spectrum, arguments about \emph{redshirting} rely mostly on anecdotal evidence (61\% of \emph{verifiable experiential} propositions). We will discuss the phenomena of narratives in argumentation in more detail later in section \ref{sec:annotation.study.2.discussion}.
All the topics except \emph{private vs. public schools} exhibit similar amount of \emph{verifiable non-experiential} premises (9\%--22\%), usually referring to expert studies or facts. However, this type of premises has usually the lowest frequency.

\begin{table}
\begin{small}
\begin{tabular}{p{1.4em}p{3.7cm}p{3.7cm}p{3.7cm}}
	&  Unverifiable	&  Verifiable non-experiential	&  Verifiable experiential	\\ \hline
\textbf{PIS}	&  73\%	& 22\%	& 4\%	\\
(112)	&  \tableex{Religion is basically a gang mentality where people feel they need to belong to a group...} \newline \tableex{A primary purpose of public education is to shape good citizens.}		&  \tableex{Fact: Muslims pray five times daily, in a way which is not practical in a normal classroom setting.} \newline \tableex{Japan, where no one prays at school, has the lowest crime rate of any developed nation.}	& \tableex{When I was a kid we learned religion in church, math, reading, history, etc., in school and at home.} \newline \tableex{I am a victim of this latter possibility. Believe me, I'm still trying to repair the damage.}\\ \hline
\textbf{HS}	&  57\%	& 14\%	& 29\% \\
(160)	& \tableex{But when you put 30 kids in one classroom, it becomes very difficult to teach them all individually.} \newline \tableex{The trouble is, home schooling can be a cover for all sorts of undesirable stuff.}	& \tableex{Only a fortnight ago a report was published by Robin Alexander and his team at Cambridge University which found that the primary school curriculum is too narrow and involves too much testing.} \newline \tableex{Dr. Smedley believes that homeschoolers have superior socialization skills, and his research supports this claim.}	& \tableex{It was boring, tedious, slow and frustrating. I learned nothing that I did not know before other than a handful of French verbs which have so far been of as much use as a chocolate fireguard.} \newline \tableex{Everyone I know went to public school and on to college. We didn't feel unprepared} \\ \hline
\textbf{SSE}	&  46\%	& 18\%	& 36\%\\
(96)	&  \tableex{Co-ed schools don't cultivate the cooperation or better understanding between the opposite sexes.} \newline  \tableex{Research is quite clear about this.}	& \tableex{Studies show that women suffer from a stereotype threat in math and science, meaning that in the fields of math and science women are more apprehensive to perform [...]} \newline  \tableex{Studies clearly establish that single-sex schools are, IN GENERAL, better for educational outcomes than co-ed schools.}	& \tableex{The unhealthiest social situations I was ever in were an all-boys school and military training.} \newline \tableex{Once I switched schools, I found myself with valuable ? and what I?m sure to be life long ? friendships with the boys I sat with in class.} \\ \hline
\textbf{MS}	&  47\%	& 10\%	& 43\%\\
(51)	&  \tableex{School and classroom design has not evolved and this has stagnated the inclusion movement.}	\newline \tableex{The level of differentiated instruction required to develop some functional skills is not possible in mainstream classrooms.} & \tableex{In a TEACH magazine article about his new book, Adelman says inclusion of students with disabilities benefits entire student bodies by teaching kids about diversity [...]} \newline \tableex{The reality is students with special needs are a small percentage of the population and cannot drive a fundamental shift in education.}	& \tableex{I have a HF autistic son w/ severe ADHD.. He is doing awesome in grade 1 he has a 1 on 1 aide in the class.. We feel well supported by the school system and he has only 18 in his class!} \newline \tableex{I often spent 20 mins of each year 9 lesson getting the boys to stop aggravating the ADHD boy, as he would then "blow", much to the amusement of everyone.}	\\ \hline
\textbf{PPS}	&  43\%	& 0\%	& 57\%\\
(159)	&  \tableex{Public schools are not about education; they are about social engineering.} \newline \tableex{Kids are indoctrinated not educated in Public Schools.}	& ---	& \tableex{Worked for us.} \newline \tableex{I have five Daughters and they all went to private schools and everyone of them have a degree and now have good paying jobs.}	\\ \hline
\textbf{RS}	&  30\%	& 9\%	& 61\%\\
(67)	& \tableex{they will grow up,, they will mature.} \newline \tableex{These kids need to be prepared for the 21st centuary global economy by being enrolled in a local second language immersion kindergarden as soon as they can enroll.}	& \tableex{There have been a lot of studies that show people born later in the year (March) are more successful in life due to the age gap.} \newline \tableex{Studies show the practice (when common) has socioeconomic repurcissions down the line and can increase the HS drop out rate  [...]} & \tableex{I honestly made my decision because I had a choice and I just did not feel right personally about sending my DS to Kindergarten at age 4.} \newline \tableex{However with my oldest if he were born a couple of months earlier I would have kept him back a year as he would not have been ready.} \\
\end{tabular}
\end{small}
\caption{\label{tab:support} Distribution of premise types for each topic with examples. \textbf{HS} -- homeschooling, \textbf{MS} -- mainstreaming, \textbf{PIS} -- prayer in schools, \textbf{PPS} -- private vs. public schools,  \textbf{RS} -- redshirting,  \textbf{SSE} -- single sex education. Number of analyzed premises shown in parentheses.}
\end{table}

\subsubsection{Discussion}
\label{sec:annotation.study.2.discussion}

Manually analyzing argumentative discourse and reconstructing (annotating) the underlying argument structure and its components is difficult.
As \namecite[p.~267]{Reed2006} point out, \emph{``the analysis of arguments is often hard, not only for students, but for experts too.''}
According to \namecite[p.~81]{Harrell.2011b}, argumentation is a skill and \emph{``even for simple arguments, untrained college students can identify the conclusion but without prompting are poor at both identifying the premises and how the premises support the conclusion.''}
\namecite[p.~81]{Harrell.2011} further claims that \emph{``a wide literature supports the contention that the particular skills of understanding, evaluating, and producing arguments are generally poor in the population of people who have not had specific training and that specific training is what improves these skills.''} Some studies, for example, show that students perform significantly better on reasoning tasks when they have learned to identify premises and conclusions \cite{Shaw.1996} or have learned some standard argumentation norms \cite{Weinstock.et.al.2004}.

One particular extra challenge in analyzing argumentation in Web user-generated discourse is that the authors produce their texts probably without any existing argumentation theory or model in mind.\footnote{By analyzing the arguments during annotation, our impression was that an average user participating in on-line discussions is not a skilled arguer. However, we miss grounded empirical evidence to support such a claim.}
We assume that argumentation or persuasion is inherent when users discuss controversial topics, but the true reasons why people participate in on-line communities and what drives their behavior is another research question \cite{Bishop.2007,Sun.et.al.2014,deMeloBezerra.Hirata.2011,Cullen.Morse.2011}. When the analyzed texts have a clear intention to produce argumentative discourse, such as in argumentative essays \cite{Stab.Gurevych.2014}, the argumentation is much more explicit and a substantially higher inter-annotator agreement can be achieved.

\paragraph{Suitability of the modified Toulmin's model}

The model seems to be suitable for short persuasive documents, such as comments and forum posts. Its applicability to longer documents, such as articles or blog posts, is problematic for several reasons.

The argument components of the (modified) Toulmin's model and their roles are not expressive enough to capture argumentation that not only conveys the logical structure (in terms of reasons put forward to support the claim), but also relies heavily on the rhetorical power. This involves various stylistic devices, pervading narratives, direct and indirect speech, or interviews.\footnote{For a deep analysis of the role of direct speech in newspaper discourse argumentation, see \cite{Smirnova.2009}.} While in some cases the argument components are easily recognizable, the vast majority of the discourse in articles and blog posts does not correspond to any distinguishable argumentative function in the \emph{logos} dimension. As the purpose of such discourse relates more to rhetoric than to argumentation, unambiguous analysis of such phenomena goes beyond capabilities of the current argumentation model. For a discussion about metaphors in Toulmin's model of argumentation see, e.g., \cite{Xu.Wu.2014,Santibanez.2010}.

Articles without a clear standpoint towards the discussed controversy cannot be easily annotated with the model either. Although the matter is viewed from both sides and there might be reasons presented for either of them, the overall persuasive intention is missing and fitting such data to the argumentation framework causes disagreements.\footnote{Note that we only filtered persuasive documents in annotation study 1 (section \ref{sec:annotation.study.1}) for comments and forum posts; blog posts and newswire articles were checked only briefly while collecting the \emph{raw corpus}.} One solution might be to break the document down to paragraphs and annotate each paragraph separately, examining argumentation on a different level of granularity.

\paragraph{Annotating other dimensions of argument}
\label{sec:annotating.pathos}

As introduced in section \ref{sec:dimensions.of.argument}, there are several dimensions of an argument. The Toulmin's model focuses solely on the \emph{logos} dimension. We decided to ignore the \emph{ethos} dimension, because dealing with the author's credibility remains unclear, given the variety of the source web data.\footnote{Modeling influential persons belongs to research in social network analysis, which is beyond the scope of this article.}
However, exploiting the \emph{pathos} dimension of an argument is prevalent in the web data, for example as an appeal to emotions. Therefore we experimented with annotating \emph{appeal to emotions} as a separate category independent of components in the \emph{logos} dimension.
We defined some features for the annotators how to distinguish \emph{appeal to emotions}. Figurative language such as hyperbole, sarcasm, or obvious exaggerating to ``spice up'' the argument are the typical signs of pathos. In an extreme case, the whole argument might be purely emotional, as in the following example.

\begin{quote}
\textbf{Doc\#1698 (comment, prayer in schools)} 
\appealtoemotion{Prayer being removed from school is just the leading indicator of a nation that is ‘Falling Away’ from Jehovah. [...] And the disasters we see today are simply God’s finger writing on the wall: Mene, mene, Tekel, Upharsin; that is, God has weighed America in the balances, and we’ve been found wanting. No wonder 50 million babies have been aborted since 1973. [...]}
\end{quote}

We kept annotations on the \emph{pathos} dimension as simple as possible (with only one \emph{appeal to emotions} label), but the resulting agreement was unsatisfying ($\alphaunit$ 0.30) even after several annotation iterations.
Appeal to emotions is considered as a type of fallacy \cite{Govier.2010,Damer.2013}. Given the results, we assume that more carefully designed approach to fallacy annotation should be applied.  To the best of our knowledge, there have been very few research works on modeling fallacies similarly to arguments on the discourse level \cite{Pineau2013a}. Therefore the question, in which detail and structure fallacies should be annotated, remains open. For the rest of the paper, we thus focus on the \emph{logos} dimension solely.

\paragraph{Narratives in argumentation}
\label{par:narratives.in.argumentation}

Some of the educational topics under examination relate to young children (e.g., redshirting or mainstreaming); therefore we assume that the majority of participants in discussions are their parents. We observed that many documents related to these topics contain narratives. Sometimes the story telling is meant as a support for the argument, but there are documents where the narrative has no intention to persuade and is simply a story sharing.

There is no widely accepted theory of the role of narratives among argumentation scholars. According to \namecite{Fisher.1987}, humans are storytellers by nature, and the ``reason'' in argumentation is therefore better understood in and through the narratives. He found that good reasons often take the form of narratives. \namecite{Hoeken.Fikkers.2014} investigated how integration of explicit argumentative content into narratives influences issue-relevant thinking and concluded that identifying with the character being in favor of the issue yielded a more positive attitude toward the issue. In a recent research, \namecite{Bex.2011} proposes an argumentative-narrative model of reasoning with evidence, further elaborated in \cite{Bex.et.al..2012}; also \namecite{Niehaus.et.al.2012} proposes a computational model of narrative persuasion.

Stemming from another research field, \namecite{LeytonEscobar2014} found that online community members who use and share narratives have higher participation levels and that narratives are useful tools to build cohesive cultures and increase participation. \namecite{Betsch.et.al.2010} examined influencing vaccine intentions among parents and found that narratives carry more weight than statistics.

\subsection{Summary of annotation studies}

This section described two annotation studies that deal with argumentation in user-generated Web content on different levels of detail. In section \ref{sec:annotation.study.1}, we argued for a need of document-level distinction of persuasiveness. We annotated 990 comments and forum posts, reaching moderate inter-annotator agreement (Fleiss' $\pi$ 0.59). Section \ref{sec:annotation.study.2} motivated the selection of a model for micro-level argument annotation, proposed its extension based on pre-study observations, and outlined the annotation set-up. This annotation study resulted into 340 documents annotated with the modified Toulmin's model and reached moderate inter-annotator agreement in the logos dimension (Krippendorff's $\alphaunit$ 0.48). These results make the annotated corpora suitable for training and evaluation computational models and each of these two annotation studies will have their experimental counterparts in the following section.

\section{Experiments}
\label{sec:experiments}

This section presents experiments conducted on the annotated corpora introduced in section \ref{sec:annotation.studies}. We put the main focus on identifying \emph{argument components} in the discourse.\footnote{
We also experimented with classification of \emph{persuasive} documents, as introduced in the annotation study 1 (section \ref{sec:annotation.study.1}). 
This task can be seen as standard document-level two-class text classification.
Using SVM \cite{Cortes.Vapnik.1995} with Sequential Minimal Optimization \cite{Platt.1999}, polynomial kernel, and $n-$gram baseline features, we obtained 0.69 Macro $F_1$ score.
We also employed a rich feature set (a large part of features that will be discussed in section \ref{sec:identification.of.argument.components}) but the system did not beat the baseline, therefore we do not report on this experiment in detail.
However, we expect that in a real-world scenario of automatically analyzing argument components in user-generated content, the first step of assessing on-topic persuasiveness (or external relevance \cite{Paglieri.Castelfranchia.2014}) is essential.
}
To comply with the machine learning terminology, in this section we will use the term \emph{domain} as an equivalent to a topic (remember that our dataset includes six different topics; see section \ref{sec:topics.and.registers}).

We evaluate three different scenarios. First, we report \emph{ten-fold cross validation} over a random ordering of the entire data set. Second, we deal with \emph{in-domain ten-fold cross validation} for each of the six domains. Third, in order to evaluate the domain portability of our approach, we train the system on five domains and test on the remaining one for all six domains (which we report as \emph{cross-domain validation}).

\subsection{Identification of argument components}
\label{sec:identification.of.argument.components}

In the following experiment, we focus on automatic identification of arguments in the discourse. Our approach is based on supervised and semi-supervised machine learning methods on the \emph{gold data Toulmin} dataset introduced in section \ref{sec:annotation.study.2}.

An argument consists of different components (such as \emph{premises}, \emph{backing}, etc.) which are implicitly linked to the \emph{claim}. In principle one document can contain multiple independent arguments.\footnote{In our approach to annotation of controversies, this would mean that the \emph{overall} standpoint of the author is neutral but she presents arguments for both sides of the controversy.} However, only 4\% of the documents in our dataset contain arguments for both sides of the issue. Thus we simplify the task and assume there is only one argument per document.\footnote{This simplification can be seen as a limitation of our model, as argumentation mining in some related works is a form of structured predictions of elements in discourse where the explicit notion of relation between argument components is crucial for argument `parsing', e.g., in the work by \namecite{Peldszus.Stede.EMNLP.2015}, envisioned in their earlier survey paper \cite{Peldszus2013a}, or by \namecite{Stab.Gurevych.2014b}. It is thus possible that in a general argumentative discourse, the same proposition can play two different roles in two arguments, similarly to the approach of \namecite{Aharoni.et.al.2014}. This phenomena was discussed as \emph{divergent structures} by \namecite{Thomas.1981} and later elaborated on by \namecite[p.~16]{Freeman.2011}.}

Given the low inter-annotator agreement on the \emph{pathos} dimension (Table \ref{tab:iaa.study2}), we focus solely on recognizing the logical dimension of argument. The \emph{pathos} dimension of argument remains an open problem for a proper modeling as well as its later recognition.

\subsubsection{Data representation and evaluation}
\label{sec:data.representation.evaluation}

Since the smallest annotation unit is a token and the argument components do not overlap, we approach identification of argument components as a sequence labeling problem. We use the BIO encoding, so each token belongs to one of the following 11 classes: \emph{O} (not a part of any argument component), \emph{Backing-B}, \emph{Backing-I}, \emph{Claim-B}, \emph{Claim-I}, \emph{Premise-B}, \emph{Premise-I}, \emph{Rebuttal-B}, \emph{Rebuttal-I}, \emph{Refutation-B}, \emph{Refutation-I}. This is the minimal encoding that is able to distinguish two adjacent argument components of the same type. In our data, 48\% of all adjacent argument components of the same type are direct neighbors (there are no "O" tokens in between).

We report Macro-$F_1$ score and $F_1$ scores for each of the 11 classes as the main evaluation metric. This evaluation is performed on the token level, and for each token the predicted label must exactly match the gold data label (classification of tokens into 11 classes).

As instances for the sequence labeling model, we chose \emph{sentences} rather than \emph{tokens}. During our initial experiments, we observed that building a sequence labeling model for recognizing argument components as sequences of \emph{tokens} is too fine-grained, as a single token does not convey enough information that could be encoded as features for a machine learner.
However, as discussed in section \ref{sec:annotation-set-up}, the annotations were performed on data pre-segmented to sentences and annotating tokens was necessary only when the sentence segmentation was wrong or one sentence contained multiple argument components.
Our corpus consists of 3899 sentences, from which 2214 sentences (57\%) contain no argument component. From the remaining ones, only 50 sentences (1\%) have more than one argument component. Although in 19 cases (0.5\%) the sentence contains a \emph{Claim}-\emph{Premise} pair which is an important distinction from the argumentation perspective, given the overall small number of such occurrences, we simplify the task by treating each sentence as if it has either one argument component or none.

\begin{figure}
\begin{center}
\includegraphics[width=0.9\textwidth]{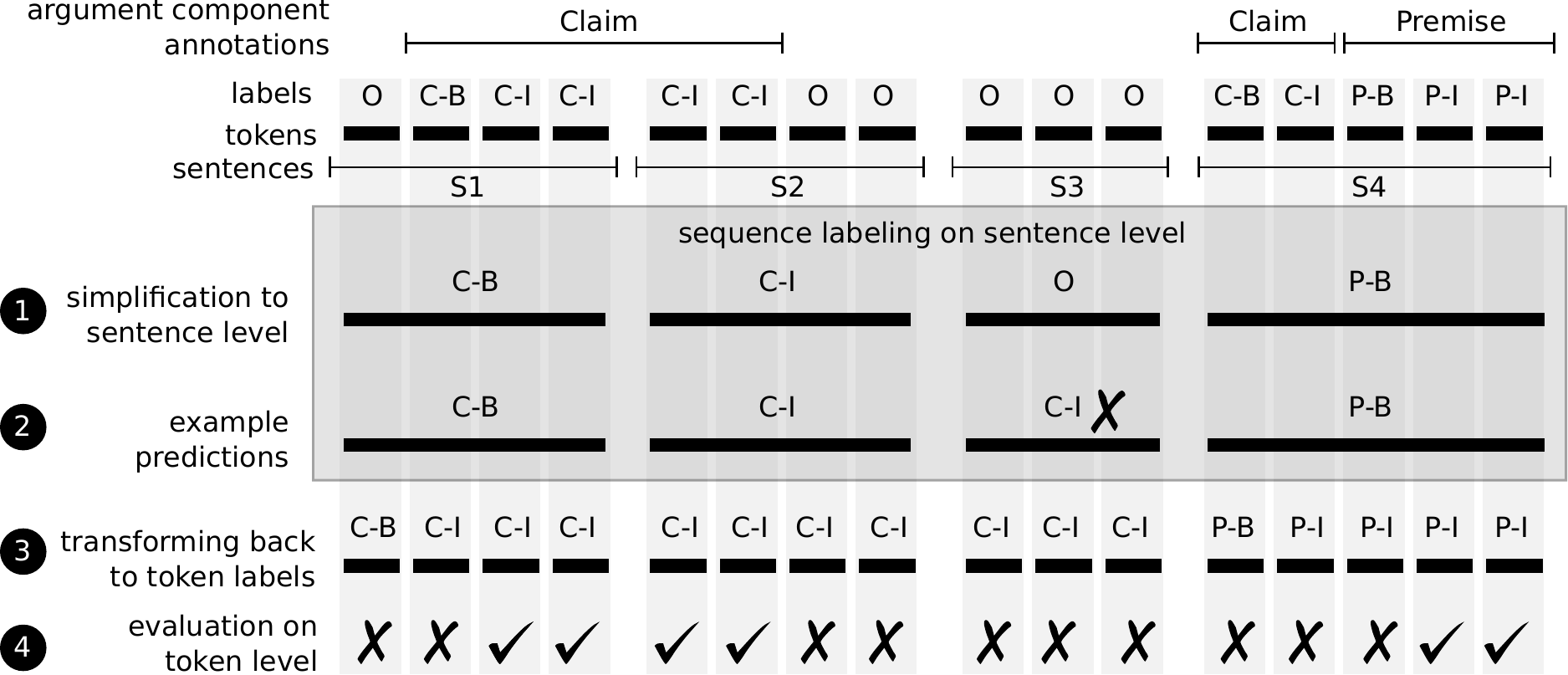}
\caption{\label{fig:segmentation-approximation} Our approach to simplifying argument component segmentation and evaluation of the system. Gold data are labeled on the token level (\emph{C} = \emph{Claim}, \emph{P} = \emph{Premise}). In step 1, the argument component label becomes a new label for the entire sentence. The resulting label reflects if the component begins in the sentence (i.e., the case of \emph{Claim-B} in \emph{S1}). If there are more components in one sentence, the longest one is selected (i.e., the case of \emph{Claim} and \emph{Premise} in \emph{S4}). In step 2, the predictions are obtained for entire sentences, as one sentence represents the minimal unit in sequence labeling mode. In step 3, the labels are translated back to the token level, spanning over the entire sentence. If the predicted label is a \emph{*-B} tag, the first token is labeled as \emph{*-B} and the remaining ones as \emph{*-I} (i.e., in \emph{S1} and \emph{S4}). In step 4, evaluation of predictions is performed solely on the token level by comparing the predicted token labels with the gold token labels.}
\end{center}
\end{figure}

The approximation with sentence-level units is explained in the example in Figure \ref{fig:segmentation-approximation}. In order to evaluate the expected performance loss using this approximation, we used an \emph{oracle} that always predicts the correct label for the unit (sentence) and evaluated it against the true labels (recall that the evaluation against the true gold labels is done always on token level). We lose only about 10\% of macro $F_1$ score (0.906) and only about 2\% of accuracy (0.984). This performance is still acceptable, while allowing to model sequences where the minimal unit is a sentence.

\subsubsection{Gold data statistics}

Table \ref{tab:gold.data.distribution.task2} shows the distribution of the classes in the \emph{gold data Toulmin}, where the labeling was already mapped to the sentences. The little presence of \emph{rebuttal} and \emph{refutation} (4 classes account only for 3.4\% of the data) makes this dataset very unbalanced.

\begin{table}
\begin{tabular}{lrr|lrr}
& \multicolumn{2}{c}{\textbf{Sentences in data}} & & \multicolumn{2}{c}{\textbf{Sentences in data}} \\
\textbf{Class} & \textbf{Relative (\%)} & \textbf{Absolute} & 
\textbf{Class} & \textbf{Relative (\%)} & \textbf{Absolute} \\ \hline
Backing-B &     5.6  &  220 & Premise-I &    8.6 &  336 \\
Backing-I &    7.2  & 281 & Rebuttal-B  &  1.6 &  61 \\
Claim-B   &    4.4 &  171 & Rebuttal-I  &  0.9 &  37 \\
Claim-I   &    0.4 &  16 & Refutation-B & 0.5 &  18 \\
O         &    56.8  & 2214 & Refutation-I & 0.4  & 15 \\
Premise-B &    13.6 & 530 & \textbf{Total} & & 3899 \\
\end{tabular}
\caption{\label{tab:gold.data.distribution.task2} Class distribution of the \emph{gold data Toulmin} corpus approximated to the sentence level boundaries.}
\end{table}

\subsubsection{Methods and features}
\label{sec:methods.and.features}

We chose \svmhmm \cite{Joachims.et.al.2009} implementation\footnote{\url{http://www.cs.cornell.edu/people/tj/svm_light/svm_hmm.html}} of Structural Support Vector Machines\footnote{
Another widely used method for sequence labeling is Conditional Random Fields (CRF) \cite{Lafferty.et.al.2001}, but the performance of CRF has been found comparable to \svmhmm by \namecite{Keerthi.Sundararajan.2007}.
} for sequence labeling.\footnote{
Argument components can span several sentences, their boundaries are not fixed. Therefore, each sentence belonging to a particular argument component can be encoded with two different tags, namely the begin tag (i.e., \emph{Claim-B}) for the first sentence and the ``in'' tag (i.e., \emph{Claim-I}) for the following sentences. Although this can be treated as a simple sentence classification task, sequence labeling can leverage the probability distribution of label sequences (for instance \emph{Claim-B}, \emph{Claim-I} are more likely to occur then \emph{Claim-B}, \emph{Premise-I}).
}
Each sentence ($\mathbf{x}$) is represented as a vector of real-valued features.

We defined the following feature sets:

\begin{itemize}
\item \textbf{FS0}: Baseline lexical features
	\begin{itemize}
	\item word uni-, bi-, and tri-grams (binary)
	\end{itemize}

\item \textbf{FS1}: Structural, morphological, and syntactic features
	\begin{itemize}
	\item First and last 3 tokens. \emph{Motivation:} these tokens may contain discourse markers or other indicators for argument components, such as ``therefore'' and ``since'' for premises or ``think'' and ``believe'' for claims.
	\item Relative position in paragraph and relative position in document. \emph{Motivation:} We expect that claims are more likely to appear at the beginning or at the end of the document.
	\item Number of POS 1-3 grams, dependency tree depth, constituency tree production rules, and number of sub-clauses. Based on \cite{Stab.Gurevych.2014b}.
	\end{itemize}
\item \textbf{FS2}: Topic and sentiment features
	\begin{itemize}
	\item 30 features taken from a vector representation of the sentence obtained by using Gibbs sampling on LDA model \cite{Blei.et.al.2003,McCallum.2002} with 30 topics trained on unlabeled data from the \emph{raw corpus}. \emph{Motivation:} Topic representation of a sentence might be valuable for detecting off-topic sentences, namely non-argument components.
	\item  Scores for five sentiment categories (from very negative to very positive) obtained from Stanford sentiment analyzer \cite{Socher.et.al.2013}. \emph{Motivation:} Claims usually express opinions and carry sentiment.
	\end{itemize}
\item \textbf{FS3}: Semantic, coreference, and discourse features
	\begin{itemize}
	\item Binary features from Clear NLP Semantic Role Labeler \cite{Choi.2012}. Namely, we extract \emph{agent}, \emph{predicate + agent}, \emph{predicate + agent + patient + (optional) negation}, \emph{argument type + argument value}, and \emph{discourse marker} which are based on PropBank semantic role labels.\footnote{Explained in detail in annotation guidelines at \url{http://clear.colorado.edu/compsem/documents/propbank_guidelines.pdf}} \emph{Motivation:} Exploit the semantics of Capturing the semantics of the sentences.
	\item Binary features from Stanford Coreference Chain Resolver \cite{Lee.et.al.2013}, e.g., presence of the sentence in a chain, transition type (i.e., nominal--pronominal), distance to previous/next sentences in the chain, or number of inter-sentence coreference links. \emph{Motivation:} Presence of coreference chains indicates links outside the sentence and thus may be informative, for example, for classifying whether the sentence is a part of a larger argument component.
	\item Results of a PTDB-style discourse parser \cite{Lin.et.aol2014}, namely the type of discourse relation (explicit, implicit), presence of discourse connectives, and attributions. \emph{Motivation:} It has been claimed that discourse relations play a role in argumentation mining \cite{Cabrio2013b}.
	\end{itemize}
\item \textbf{FS4}: Embedding features
	\begin{itemize}
	\item 300 features from word embedding vectors using word embeddings trained on part of the Google News dataset \cite{Mikolov.et.al.2013}. In particular, we sum up embedding vectors (dimensionality 300) of each word, resulting into a single vector for the entire sentence. This vector is then directly used as a feature vector.\footnote{\namecite{Le.Mikolov.2014} proposed more advanced techniques for sentence representation using embeddings.} \emph{Motivation:} Embeddings helped to achieve state-of-the-art results in various NLP tasks \cite{Socher.et.al.2013,Guo.et.al.2014}.
	\end{itemize}
\end{itemize}

Except the baseline lexical features, all feature types are extracted not only for the current sentence $s_i$, but also for $C$ preceding and subsequent sentences, namely $s_{i - C}$, $s_{i -  C + 1}$, $\dots$ $s_{i + C - 1}$, $s_{i + C}$, where $C$ was empirically set to 4.\footnote{We used grid search with different $C$ values in several feature set combinations (FS01, FS012) over the entire cross-validation scenario and fixed the value afterwards.} Each feature is then represented with a prefix to determine its relative position to the current sequence unit.\footnote{For example, \texttt{minus2Sent\_sentimentNegative=0.23} or \texttt{plus1Sent\_DependencyTreeDepth=3}.}

\subsubsection{Results}

\begin{table}
\begin{small}
\begin{tabular}{lr|rr|rrrrrrrrr}
\multicolumn{4}{l}{ } & \multicolumn{9}{c}{\textbf{Feature set combinations}} \\
& \textbf{Hum}	& \textbf{Ran}	&  \textbf{"O"}	& 0	& 01	& 012	& 0123	& 01234	& 1234	& 234	& 34	& 4\\ \hline
M-$F_1$	& .602	& .071	& .065	& .156	& .217	& .229	& .219	& \textbf{.251}	& .240	& \textbf{.251}	& .238	& .229\\ \hline
Bac-B	& .664	& .063	& .000	& .140	& .262	& .311	& .294	& .320	& \textbf{.326}	& .291	& .278	& .278\\
Bac-I	& .579	& .120	& .000	& .159	& .339	& .364	& .334	& .372	& \textbf{.380}	& .366	& .362	& .363\\
Cla-B	& .739	& .051	& .000	& .127	& .203	& .234	& .211	& .257	& .259	& \textbf{.270}	& .266	& .252\\
Cla-I	& .728	& .051	& .000	& .165	& .207	& .224	& .194	& .237	& .245	& \textbf{.269}	& .258	& .242\\
O	& .833	& .136	& .707	& \textbf{.714}	& .705	& .703	& .708	& .691	& .686	& .675	& .671	& .669\\
Pre-B	& .673	& .082	& .000	& .176	& .280	& .289	& .286	& \textbf{.298}	& .294	& .265	& .269	& .246\\
Pre-I	& .736	& .179	& .000	& .241	& .390	& .391	& .380	& \textbf{.400}	& .396	& .357	& .366	& .356\\
Reb-B	& .403	& .026	& .000	& .000	& .000	& .000	& .000	& .082	& .000	& \textbf{.037}	& .000	& .035\\
Reb-I	& .495	& .053	& .000	& .000	& .000	& .000	& .000	& .104	& .054	& \textbf{.118}	& .036	& .076\\
Ref-B	& .390	& .000	& .000	& .000	& .000	& .000	& .000	& .000	& .000	& \textbf{.057}	& \textbf{.057}	& .000\\
Ref-I	& .387	& .023	& .000	& .000	& .000	& .000	& .000	& .000	& .000	& \textbf{.055}	& .052	& .000\\
\end{tabular}
\end{small}
\caption{\label{tab:results.task2.1} 10-fold cross validation results of classification of argument components using different feature sets. Macro-$F_1$ and $F_1$ scores for individual classes are shown. Column \textbf{Hum} denotes human performance, \textbf{Ran} is a random classifier, \textbf{"O"} is a majority voting (each token is labeled as "O"). Bold numbers denote the best results for the given class. The best performing configuration is the \textbf{234} feature set. Differences between the best feature sets (01234 and 234) and other sets are statistically significant ($p < 0.001$, paired exact Liddell's test).}
\end{table}

Let us first discuss the upper bounds of the system. Performance of the three human annotators is shown in the first column of Table \ref{tab:results.task2.1} (results are obtained from a cumulative confusion matrix). The overall Macro-$F_1$ score is 0.602 (accuracy 0.754). If we look closer at the different argument components, we observe that humans are good at predicting \emph{claims}, \emph{premises}, \emph{backing} and non-argumentative text (about 0.60-0.80 $F_1$), but on \emph{rebuttal} and \emph{refutation} they achieve rather low scores. Without these two components, the overall human Macro-$F_1$ would be 0.707. This trend follows the inter-annotator agreement scores, as discussed in section \ref{sec:toulmin.IAA}.

In our experiments, the feature sets were combined in the bottom-up manner, starting with the simple lexical features (FS0), adding structural and syntactic features (FS1), then adding topic and sentiment features (FS2), then features reflecting the discourse structure (FS3), and finally enriched with completely unsupervised latent vector space representation (FS4). In addition, we were gradually removing the simple features (e.g., without lexical features, without syntactic features, etc.) to test the system with more ``abstract'' feature sets (feature ablation). The results are shown in Table \ref{tab:results.task2.1}.

The overall best performance (Macro-$F_1$ 0.251) was achieved using the rich feature sets (01234 and 234) and significantly outperformed the baseline as well as other feature sets. Classification of non-argumentative text (the "O" class) yields about 0.7 $F_1$ score even in the baseline setting. The boundaries of \emph{claims} (\textbf{Cla-B}), \emph{premises} (\textbf{Pre-B}), and \emph{backing} (\textbf{Bac-B}) reach in average lower scores then their respective inside tags (\textbf{Cla-I}, \textbf{Pre-I}, \textbf{Bac-I}). It can be interpreted such that the system is able to classify that a certain sentence belongs to a certain argument component, but the distinction whether it is a beginning of the argument component is harder. The very low numbers for \emph{rebuttal} and \emph{refutation} have two reasons. First, these two argument components caused many disagreements in the annotations, as discussed in section \ref{sec:iaa.toulmin.causes.of.disagreement.qualitative}, and were hard to recognize for the humans too. Second, these four classes have very few instances in the corpus (about 3.4\%, see Table \ref{tab:gold.data.distribution.task2}), so the classifier suffers from the lack of training data.

\begin{table}
\begin{tabular}{l|rrrrrrrrr}
 & \multicolumn{9}{c}{\textbf{Feature set combinations}} \\
Domain & 0	& 01	& 012	& 0123	& 01234	& 1234	& 234	& 34	& 4\\ \hline
\textbf{HS}	&0.134	&0.162	&0.167	&0.165	&0.187	&0.176	&\textbf{0.205}	&0.203	&0.193\\
\textbf{MS}	&0.072	&0.123	&0.138	&0.151	&0.198	&0.216	&0.165	&0.190	&\textbf{0.226}\\
\textbf{PIS}	&0.152	&0.174	&0.178	&0.168	&\textbf{0.212}	&0.192	&0.175	&0.177	&0.181\\
\textbf{PPS}	&0.235	&0.233	&0.230	&0.240	&\textbf{0.265}	&0.250	&0.239	&0.250	&0.243\\
\textbf{RS}	&0.090	&0.156	&0.156	&0.144	&0.195	&0.201	&0.204	&0.190	&\textbf{0.225}\\
\textbf{SSE}	&0.141	&0.176	&0.200	&0.185	&0.206	&\textbf{0.216}	&0.189	&0.202	&0.201\\ \hline
\textbf{Aggregated} & 0.182 & 0.200 & 0.205 & 0.206 & \textbf{0.236} & 0.230 & 0.218 & 0.228 & 0.229 \\
\end{tabular}
\caption{\label{tab:results.task2.2-in} Results of classification of argument components in the in-domain cross-validation scenario. Macro-$F_1$ scores reported, bold numbers denote the best results. \textbf{HS} -- homeschooling, \textbf{MS} -- mainstreaming, \textbf{PIS} -- prayer in schools, \textbf{PPS} -- private vs. public schools,  \textbf{RS} -- redshirting,  \textbf{SSE} -- single sex education. Results in the \emph{aggregated} row are computed from an aggregated confusion matrix over all domains. The differences between the best feature set combination (01234) and others are statistically significant ($p < 0.001$; paired exact Liddell's test).}
\end{table}

\begin{table}
\begin{tabular}{l|rrrrrrrrr}
 & \multicolumn{9}{c}{\textbf{Feature set combinations}} \\
Domain & 0	& 01	& 012	& 0123	& 01234	& 1234	& 234	& 34	& 4\\ \hline
\textbf{HS}	&0.087	&0.063	&0.044	&0.106	&0.072	&0.075	&0.065	&0.063	&\textbf{0.197}\\
\textbf{MS}	&0.072	&0.060	&0.070	&0.058	&0.038	&0.062	&0.045	&0.060	&\textbf{0.188}\\
\textbf{PIS}	&0.078	&0.073	&0.083	&0.074	&0.086	&0.073	&0.096	&0.081	&\textbf{0.166}\\
\textbf{PPS}	&0.070	&0.059	&0.070	&0.132	&0.059	&0.062	&0.071	&0.067	&\textbf{0.203}\\
\textbf{RS}	&0.067	&0.067	&0.082	&0.110	&0.097	&0.092	&0.075	&0.075	&\textbf{0.257}\\
\textbf{SSE}	&0.092	&0.089	&0.066	&0.036	&0.120	&0.091	&0.071	&0.066	&\textbf{0.194}\\ \hline
\textbf{Aggregated} & 0.079 & 0.086 & 0.072 & 0.122 & 0.094 & 0.088 & 0.089 & 0.076 & \textbf{0.209}
\end{tabular}
\caption{\label{tab:results.task2.2-cross} Results of classification of argument components in the cross-domain scenario. Macro-$F_1$ scores reported, bold numbers denote the best results. \textbf{HS} -- homeschooling, \textbf{MS} -- mainstreaming, \textbf{PIS} -- prayer in schools, \textbf{PPS} -- private vs. public schools,  \textbf{RS} -- redshirting,  \textbf{SSE} -- single sex education. Results in the \emph{aggregated} row are computed from an aggregated confusion matrix over all domains. The differences between the best feature set combination (4) and others are statistically significant ($p < 0.001$; paired exact Liddell's test).}
\end{table}

The results for the \emph{in-domain cross validation scenario} are shown in Table \ref{tab:results.task2.2-in}. Similarly to the cross-validation scenario, the overall best results were achieved using the largest feature set (01234). For \emph{mainstreaming} and \emph{red-shirting}, the best results were achieved using only the feature set 4 (embeddings). These two domains contain also fewer documents, compared to other domains (refer to Table \ref{tab:toulmin.gold.topic.register.distribution}). We suspect that embeddings-based features convey important information when not enough in-domain data are available. This observation will become apparent in the next experiment.

The \emph{cross-domain} experiments yield rather poor results for most of the feature combinations (Table \ref{tab:results.task2.2-cross}). However, using only feature set 4 (embeddings), the system performance increases rapidly, so it is even comparable to numbers achieved in the \emph{in-domain} scenario. These results indicate that embedding features generalize well across domains in our task of argument component identification. We leave investigating better performing vector representations, such as \emph{paragraph vectors} \cite{Le.Mikolov.2014}, for future work.

\begin{table}
\centering
\begin{small}
\pgfplotstabletypeset[color cells]{
x,Bac-B,Bac-I,Cla-B,Cla-I,O,Pre-B,Pre-I,Reb-B,Reb-I,Ref-B,Ref-I
Bac-B,28,7,6,0,41,12,5,0,0,0,0
Bac-I,0,35,0,3,36,0,23,0,2,0,1
Cla-B,5,3,24,4,48,10,5,1,1,0,0
Cla-I,0,6,0,26,55,0,13,0,1,0,0
O,0,8,0,2,70,0,18,0,1,0,0
Pre-B,5,4,4,1,49,24,8,2,1,0,0
Pre-I,0,9,0,3,49,0,35,0,4,0,0
Reb-B,3,7,12,4,49,18,3,3,1,0,0
Reb-I,0,9,0,6,44,0,28,0,11,0,1
Ref-B,8,0,0,12,46,8,19,0,4,4,0
Ref-I,0,4,0,4,60,0,27,0,1,0,4
}
\end{small}
\caption{\label{tab:cv-confusion-matrix} Probabilistic confusion matrix for the cross-validation scenario for the best performing system from Table \ref{tab:results.task2.1}. Row labels represent gold labels, column labels are predictions. Values in \%, each row sums up to 100\%. }
\end{table}

\subsubsection{Error analysis}

Error analysis based on the probabilistic confusion matrix \cite{Wang.et.al.2013} shown in Table \ref{tab:cv-confusion-matrix} reveals further details. About a half of the instances for each class are misclassified as non-argumentative (the "O" prediction).

\emph{Backing-B} is often confused with \emph{Premise-B} (12\%) and \emph{Backing-I} with \emph{Premise-I} (23\%). Similarly, \emph{Premise-I} is misclassified as \emph{Backing-I} in 9\%. This shows that distinguishing between \emph{backing} and \emph{premises} is not easy because these two components are similar such that they support the \emph{claim}, as discussed in section \ref{sec:iaa.toulmin.causes.of.disagreement.qualitative}. We can also see that the misclassification is consistent among \emph{*-B} and \emph{*-I} tags.

\emph{Rebuttal} is often misclassified as \emph{Premise} (28\% for \emph{Rebuttal-I} and 18\% for \emph{Rebuttal-B}; notice again the consistency in \emph{*-B} and \emph{*-I} tags). This is rather surprising, as one would expect that \emph{rebuttal} would be confused with a \emph{claim}, because its role is to provide an opposing view.

\emph{Refutation-B} and \emph{Refutation-I} is misclassified as \emph{Premise-I} in 19\% and 27\%, respectively. This finding confirms the discussion in section \ref{sec:iaa.toulmin.causes.of.disagreement.qualitative}, because the role of \emph{refutation} is highly context-dependent. In a pragmatic perspective, it is put forward to indirectly support the \emph{claim} by attacking the \emph{rebuttal}, thus having a similar function to the \emph{premise}.

\subsubsection{Qualitative error analysis}
\label{sec:qualitative.error.analysis}

We manually examined miss-classified examples produced the best-performing system to find out which phenomena pose biggest challenges. Properly detecting boundaries of argument components caused problems, as shown in Figure \ref{fig:examples-predicted} (a). This goes in line with the granularity annotation difficulties discussed in section \ref{sec:iaa.toulmin.causes.of.disagreement.qualitative}.
The next example in Figure \ref{fig:examples-predicted} (b) shows that even if boundaries of components were detected precisely, the distinction between \emph{premise} and \emph{backing} fails. The example also shows that in some cases, labeling on clause level is required (left-hand side \emph{claim} and \emph{premise}) but the approximation in the system cannot cope with this level of detail (as explained in section \ref{sec:data.representation.evaluation}).
Confusing non-argumentative text and argument components by the system is sometimes plausible, as is the case of the last rhetorical question in Figure \ref{fig:examples-predicted} (c). On the other hand, the last example in Figure \ref{fig:examples-predicted} (d) shows that some claims using figurative language were hard to be identified. The complete predictions along with the gold data are publicly available.\footnote{\url{https://www.ukp.tu-darmstadt.de/data/argumentation-mining/}}

\begin{figure*}
\begin{subfigure}{\textwidth}
\begin{footnotesize}
\begin{multicols*}{2}
\noindent\textbf{Gold} \\ 
 \noindent
Some really good points have been expressed here. [...] \newline 
\premise{
We've heard about public school space being allotted to accommodate one religion and its demand for a dedicated space. Muslim prayer is strictly segregated. Gender segregation violates our Charter of Rights and Freedoms which, under Section 15, prohibits discrimination on the grounds of race; national or ethnic origin; colour; religion; gender; age; and mental or physical disability. Sexual orientation has recently been recognized as a prohibited ground for discrimination under the Charter.} \newline 
Are gay Muslim students allowed? [...]
\columnbreak \\
\noindent \textbf{Predicted} \\ 
 \noindent
Some really good points have been expressed here. [...] \newline 
\premise{
We've heard about public school space being allotted to accommodate one religion and its demand for a dedicated space. Muslim prayer is strictly segregated.} \premise{
Gender segregation violates our Charter of Rights and Freedoms which, under Section 15, prohibits discrimination on the grounds of race; national or ethnic origin; colour; religion; gender; age; and mental or physical disability.} Sexual orientation has recently been recognized as a prohibited ground for discrimination under the Charter. \newline 
Are gay Muslim students allowed? [...]
\end{multicols*}
\end{footnotesize}
\vspace{-13pt}
\caption{\#1346 (article comment, prayer-in-schools)}
\end{subfigure}
\begin{subfigure}{\textwidth}
\begin{footnotesize}
\begin{multicols*}{2}
\noindent\textbf{Gold} \\ 
 \noindent
Ohhhh, here we go again!!!! Where ever the Muslims go, they expect "special" treatment. \claim{
No religion in schools....this is what we've come to.} \premise{
In order to keep everyone satisfied,} \claim{
there should be no religion in schools.} If parents want their children to have religion, they are going to have to teach them at home or in their places of worship.  [...] For goodness sake, it's just awful. \backing{
Time was, the school day started with a prayer, and yes, maybe religion classes, but not nowadays..there would be full scale war. If the textbooks contain any reference to God, there's trouble, and if the teachers happen to make any kind of a reference to God, or religion..or the hereafter..or what ever may have any religious connotation, the children tell their parents and the parents complain!!!!} So there you have it....yet more proof the multiculturalism doesn't work!! [...]
\columnbreak \\
\noindent \textbf{Predicted} \\ 
Ohhhh, here we go again!!!! Where ever the Muslims go, they expect "special" treatment. No religion in schools....this is what we've come to. \claim{
In order to keep everyone satisfied, there should be no religion in schools.} If parents want their children to have religion, they are going to have to teach them at home or in their places of worship. [...] For goodness sake, it's just awful. \premise{
Time was, the school day started with a prayer, and yes, maybe religion classes, but not nowadays..there would be full scale war. If the textbooks contain any reference to God, there's trouble, and if the teachers happen to make any kind of a reference to God, or religion..or the hereafter..or what ever may have any religious connotation, the children tell their parents and the parents complain!!!!} So there you have it....yet more proof the multiculturalism doesn't work!! [...]
\end{multicols*}
\end{footnotesize}
\vspace{-13pt}
\caption{\#1412 (artcomment, prayer-in-schools)}
\end{subfigure}
\begin{subfigure}{\textwidth}
\begin{footnotesize}
\begin{multicols*}{2}
\noindent\textbf{Gold} \\ 
 \noindent
\claim{
Sending your child to a private school is one of the best things you can do for them.} \premise{
The teachers do not open up a text book and teach every child the same way. Private teachers have more passion about teaching because they are free to write their own curriculum for each child based on developmental assessments and achievements to where each child is learning at the level they need to be at and not just a class as a whole.} \premise{
Children in the public school systems ARE left behind, bullied, and not challenged enough in their own learning capabilities.} When your public school system ranks \#50 in the nation you tell me, would you rather send your child to public school or honor them with attending a private school?
\columnbreak \\
\noindent \textbf{Predicted} \\ 
 \noindent
\claim{
Sending your child to a private school is one of the best things you can do for them.} The teachers do not open up a text book and teach every child the same way. Private teachers have more passion about teaching because they are free to write their own curriculum for each child based on developmental assessments and achievements to where each child is learning at the level they need to be at and not just a class as a whole. \premise{
Children in the public school systems ARE left behind, bullied, and not challenged enough in their own learning capabilities.} \claim{
When your public school system ranks \#50 in the nation you tell me, would you rather send your child to public school or honor them with attending a private school?}
\end{multicols*}
\end{footnotesize}
\vspace{-13pt}
\caption{\#2499 (artcomment, public-private-schools)}
\end{subfigure}
\begin{subfigure}{\textwidth}
\begin{footnotesize}
\begin{multicols*}{2}
\noindent\textbf{Gold} \\ 
 \noindent
\backing{
I went to both, public and private.} \premise{
The essential difference were the students. And it makes all the difference.} \claim{
The public school was a joke.}
\columnbreak \\
\noindent \textbf{Predicted} \\ 
 \noindent
\backing{
I went to both, public and private.} The essential difference were the students. And it makes all the difference. The public school was a joke.
\end{multicols*}
\end{footnotesize}
\vspace{-13pt}
\caption{\#2342 (artcomment, public-private-schools)}
\end{subfigure}
\caption{\label{fig:examples-predicted} Examples of gold data annotations on the left-hand side and system predictions in the best-performing system on the right-hand side.}
\end{figure*}

\paragraph{Hyper-parameter tuning}

\svmhmm offers many hyper-parameters with suggested default values, from which three are of importance. Parameter $t$ sets the order of dependencies of transitions in HMM, parameter $e$ sets the order of dependencies of emissions in HMM, and parameter $c$ represents a trading-off slack versus magnitude of the weight-vector.\footnote{\url{http://www.cs.cornell.edu/people/tj/svm_light/svm_hmm.html}}
For all experiments, we set all the hyper-parameters to their default values ($t = 1$, $e = 0$, $c = 5.0$). Using the best performing feature set from Table \ref{tab:results.task2.1}, we experimented with a grid search over different values ($c \in \{0.1, 1.0, 5.0, 10.0, 50.0\}$, $e \in \{0, 1\}$, $t \in \{1, 2, 3\}$) but the results did not outperform the system trained with default parameter values.

\subsubsection{Discussion}

The $F_1$ scores might seem very low at the first glance. One obvious reason is the actual performance of the system, which gives a plenty of room for improvement in the future.\footnote{We also experimented with a two-step approach consisting of (1) argument component segmentation and (2) argument component classification, but the performance of segmentation (about 0.7 $F_1$) was not promising.} But the main cause of low $F_1$ numbers is the evaluation measure --- using 11 classes on the token level is very strict, as it penalizes a mismatch in argument component boundaries the same way as a wrongly predicted argument component type. Therefore we also report two another evaluation metrics that help to put our results into a context.

\begin{itemize}
\item \emph{Krippendorff's $\alphaunit$} --- It was also used for evaluating inter-annotator agreement (see section \ref{sec:toulmin.IAA}).
\item \emph{Boundary similarity} \cite{Fournier2013} --- Using this metric, the problem is treated solely as a segmentation task without recognizing the argument component types.
\end{itemize}

\begin{table}
\begin{tabular}{lrrr}
& \textbf{Macro-$F_1$} & \textbf{Krippendorff's $\alphaunit$} & \textbf{Boundary similarity} \\ \hline
Human & 0.60 & 0.48* & 0.70 \\
Baseline & 0.16 & 0.11 & 0.18 \\
Best system & 0.25 & 0.30 & 0.32 \\
\end{tabular}
\caption{\label{tab:additional.metrics} Additional metrics to evaluate the performance of argument component identification applied to the results of 10-fold cross-validation scenario (Table \ref{tab:results.task2.1}). *Measured only on a subset of the data (refer to section \ref{sec:toulmin.IAA}).}
\end{table}

As shown in Table \ref{tab:additional.metrics} (the Macro-$F_1$ scores are repeated from Table \ref{tab:results.task2.1}), the best-performing system achieves 0.30 score using Krippendorf's $\alphaunit$, which is in the middle between the baseline and the human performance (0.48) but is considered as poor from the inter-annotator agreement point of view \cite{Artstein2008}. The boundary similarity metrics is not directly suitable for evaluating argument component classification, but reveals a sub-task of finding the component boundaries. The best system achieved 0.32 on this measure. \namecite{Vovk2013MT} used this measure to annotate argument spans and his annotators achieved 0.36 boundary similarity score. Human annotators in \cite{Fournier2013} reached 0.53 boundary similarity score.

The overall performance of the system is also affected by the accuracy of individual NPL tools used for extracting features. One particular problem is that the preprocessing models we rely on (POS, syntax, semantic roles, coreference, discourse; see section \ref{sec:methods.and.features}) were trained on newswire corpora, so one has to expect performance drop when applied on user-generated content. This is however a well-known issue in NLP \cite{Foster.et.al.2011,Eisenstein.2013,Baldwin.et.al.2013}.

To get an impression of the actual performance of the system on the data, we also provide the complete output of our best performing system in one PDF document together with the gold annotations in the logos dimension side by side in the accompanying software package. We believe this will help the community to see the strengths of our model as well as possible limitations of our current approaches.

\section{Conclusions}
\label{sec:conclusion}

Let us begin with summarizing answers to the research questions stated in the introduction. First, as we showed in section \ref{sec:our.model}, existing argumentation theories do offer models for capturing argumentation in user-generated content on the Web. We built upon the Toulmin's model and proposed some extensions.

Second, as compared to the negative experiences with annotating using Walton's schemes (see sections \ref{sec:argumentation.model.selection} and \ref{sec:rel.work.arg.min}), our modified Toulmin's model offers a trade-off between its expressiveness and annotation reliability. However, we found that the capabilities of the model to capture argumentation depend on the register and topic, the length of the document, and inherently on the literary devices and structures used for expressing argumentation as these properties influenced the agreement among annotators.

Third, there are aspects of online argumentation that lack their established theoretical counterparts, such as rhetorical questions, figurative language, narratives, and fallacies in general. We tried to model some of them in the pathos dimension of argument (section \ref{sec:annotating.pathos}), but no satisfying agreement was reached. Furthermore, we dealt with a step that precedes argument analysis by filtering documents given their persuasiveness with respect to the controversy.
Finally, we proposed a computational model based on machine learning for identifying argument components (section \ref{sec:identification.of.argument.components}).
In this identification task, we experimented with a wide range of linguistically motivated features and found that (1) the largest feature set (including n-grams, structural features, syntactic features, topic distribution, sentiment distribution, semantic features, coreference feaures, discourse features, and features based on word embeddings) performs best in both in-domain and all-data cross validation, while (2) features based only on word embeddings yield best results in cross-domain evaluation.

Since there is no one-size-fits-all argumentation theory to be applied to actual data on the Web, the argumentation model and an annotation scheme for argumentation mining is a function of the task requirements and the corpus properties.
Its selection should be based on the data at hand and the desired application.
Given the proposed use-case scenarios (section \ref{sec:introduction}) and the results of our annotation study (section \ref{sec:annotation.study.2}), we recommend a scheme based on Toulmin's model for short documents, such as comments or forum posts.

\paragraph{Summary of contributions}

In this article we presented our original research of argumentation mining in the user-generated Web discourse by collecting data in six controversial topics in education. We conducted an annotation study on 990 documents to filter persuasive comments and forum posts with inter-annotator agreement 0.59 Fleiss' $\kappa$
Then we annotated 340 documents (approx. 90k tokens) on the token level with a modified Toulmin's model and reached inter-annotator agreement 0.48 (Krippendorff's joint $\alphaunit$). We proposed a sequence labeling approach to identify argument components in the discourse and significantly ($p < 0.001$) outperformed the baseline (0.156) with overall macro-$F_1$ 0.251. We also found that a feature set based on word embeddings works well in a cross-domain scenario and reaches 0.209 macro$F_1$. We thoroughly examined errors made by the system and proposed future improvements.

As the argumentation mining field is still evolving, and to foster future research, we provide our annotation guidelines, the annotated data, the source codes for the experiments, as well as the results of our system for error analysis. We believe that keeping the whole process transparent will help to identify the strengths and possible shortcomings and will motivate the community to build upon our work.

\appendix

\appendixsection{Raw corpus compilation}
\label{appendix:raw.corpus.compilation}

Given the six controversial topics and four different registers introduced in section \ref{sec:topics.and.registers}, we compiled the \emph{raw corpus} semi-automatically. Websites with relevant comments to articles and discussions forums were identified manually (Google search engine) in order to maximize their relatedness to the search topic. We did not prefer any particular platform or data source. We extracted the texts automatically, however, we did some minimal data pre-selection and cleaning. If article comments formed a tree structure, we kept only the root comments, as they are most likely comment on the topic of the article, according to our observations. In discussion forum posts, we automatically removed all quotations (users usually quote the previous post to which they react).

Articles and blogs were also selected manually; we skimmed the texts quickly to check if they discuss the given topic in an argumentative manner.
Since we wanted to ensure the reliability of the extracted texts in terms of proper paragraph formatting and boiler-content removal, we extracted the texts manually.
For each document we also kept the paragraph formatting, as paragraphs play an important role in argumentative discourse \cite{McGee.2014}.

The top ten source domains from the total number of 117 unique domains are listed in Table \ref{tab:top.source.domains}. Table \ref{tab:raw-corpus-stats1} shows the document distribution with respect to the registers and topics.

\begin{table}
\begin{tabular}{lr|lr}
Domain & Docs  & Domain & Docs \\ \hline
living.msn.com & 2040 & www.theage.com.au & 196 \\
discussion.theguardian.com & 494 & www.forerunner.com & 169 \\
community.babycenter.com & 403 & www.netmums.com & 117 \\
www.washingtonpost.com & 398 & schoolsofthought.blogs.cnn.com & 117 \\
www.cbc.ca & 380 & www.greatschools.org & 89 \\
\end{tabular}
\caption{\label{tab:top.source.domains} Top 10 source domains in the \emph{raw corpus}}
\end{table}

\begin{table}
\begin{tabular}{l|r|r|r|r|r}
Topic $\backslash$ Register &  \textbf{Comment} & \textbf{Article} & \textbf{Blog} & \textbf{Forum post}  & \textbf{Total} \\ \hline
Redshirting & 237 & 10 & 10 & 178 & 435 \\
Single sex education & 237 & 10 & 10 & 76 & 333 \\
Prayer in schools & 547 & 10 & 11 & 240 & 808 \\
Homeschooling & 907 & 10 & 11 & 339 & 1267 \\
Mainstreaming & 33 & 12 & 10 & 134 & 189 \\
Public vs. private & 2235 & 10 & 10 & 157 & 2412 \\ \hline
Total & 4196 & 62 & 62 & 1124 & 5444 \\
\end{tabular}
\caption{\label{tab:raw-corpus-stats1} \emph{Raw corpus} statistics -- number of documents for particular topics and registers.}
\end{table}

\appendixsection{Examples of annotated documents from the second annotation study}
\label{app:toulmin.examples}

\begin{example}
An example of argument annotation with a re-stated claim and both dimensions (logos, pathos). With the first sentence (``Depriving your child of a basic education...'') the author appeals to emotions and uses figurative language (``child abuse,'' ``ruin whole life''). The argument is extracted under the original text. Phrases in italics summarize the content of the respective argument components produced by annotators.

\begin{quote}
\textbf{Doc\#45 (comment, homeschooling)} \appealtoemotion{Depriving your child of a basic education is a form of child abuse. It can ruin your child's whole life.}\P\newline
\claim{Home schooling should be illegal} unless \rebuttal{the parent can demonstrate that they are providing the same level of education as a public school.} There should be a core national curriculum and testing to ensure children are achieving at least a basic level of education.\P\newline
\premise{In an increasingly complex, global technological society, all people need to have a basic understanding of science, technology and local and global culture, just to be able to function and make informed decisions.}\P\newline
\claim{
I don't see any need for home schooling any child} unless \rebuttal{
the child has special needs or learning difficulties.} If public schools are under-performing, then the public education system needs to be improved. \premise{
Public education in the US seems to be a self-perpetuating disaster, with ignorant, uneducated, unqualified people on school boards deciding what children should learn.}\newline

\noindent \textbf{Claim} $\bullet$ ``Home schooling should be illegal'' $\bullet$ ``I don't see any need for home schooling any child''
\textbf{Premise} $\bullet$ \emph{Science and technology are not taught in HS} ``In an increasingly complex, global technological society, all people need to have a basic understanding of science, technology and local and global culture, just to be able to function and make informed decisions.'' $\bullet$ \emph{Public school in the US is bad} 
``Public education in the US seems to be a self-perpetuating disaster, with ignorant, uneducated, unqualified people on school boards deciding what children should learn.''
\textbf{Rebuttal} $\bullet$ \emph{HS is ok if parents demonstrate the same level of education as in schools} ``the parent can demonstrate that they are providing the same level of education as a public school.'' $\bullet$ \emph{HS can be allowed for kids with special needs} ``the child has special needs or learning difficulties.'' \textbf{Appeal to emotion} $\bullet$ ``Depriving your child of a basic education is a form of child abuse. It can ruin your child's whole life.''
\end{quote}
\end{example}

\begin{example}
This example contains annotations both in the logos and the pathos dimensions. The main support for the authors implicit claim starts with ``I personally am acquainted with four families ...'' Another reason is the social skills.

\begin{quote}
\textbf{Doc\#163 (comment, homeschooling)}
Thank you for bringing this tragedy to light. \backing{
I am a Christian, an educator, a student and a parent and I have seen too many children like the Powells. As an admissions officer, we had applicants whose "record keeping" consisted of sending boxes full of paper for our office to review as part of the application.} \premise{
If their students did get an interview, which was rare, they didn't have the social skills to survive the first round.}\P\newline
\premise{
I personlly am acquainted with four families who are home schooling their large families. All four have no intention of book-schooling their daughters past age 13 as they need to learn :homemaking skills". One of the girls, who has not been taught for two years, could be Josh Powell's twin. She is intelligent and desperate to learn, but her parents won't allow it.} \appealtoemotion{It is heartbreaking.\P\newline
That the Commonwealth of Virginia has such a rich tradition of the education of young people and allows this travesty is shameful.} All of us, no matter our religious beliefs, need to pray that the law changes before more smart children are left behind.\newline

\noindent \textbf{Claim} $\bullet$ \emph{Implicit: Against homeschooling}
\textbf{Premise} $\bullet$ \emph{HS kids lacked social skills} ``If their students did get an interview, which was rare, they didn't have the social skills to survive the first round.'' $\bullet$
\emph{I know families that HS but in fact do not teach their children at all} ``I personlly am acquainted with four families who are home schooling their large families. All four have no intention of book-schooling their daughters past age 13 as they need to learn :homemaking skills". One of the girls, who has not been taught for two years, could be Josh Powell's twin. She is intelligent and desperate to learn, but her parents won't allow it.''
\textbf{Backing} $\bullet$ \emph{Observations as an admission officer} ``I am a Christian, an educator, a student and a parent and I have seen too many children like the Powells. As an admissions officer, we had applicants whose "record keeping" consisted of sending boxes full of paper for our office to review as part of the application.'' \textbf{Appeal to emotion} $\bullet$ ``It is heartbreaking. That the Commonwealth of Virginia has such a rich tradition of the education of young people and allows this travesty is shameful.''
\end{quote}
\end{example}

\begin{example}
Notice the wrong capitalization and punctuation. This text had to be annotated on the token level, as the automatic sentence splitting could not cope with it properly.

\begin{quote}
\textbf{Doc\#2488 (comment, public-private-schools)}
BIG E.\premise{what about all the money we do send to our schools .does it help our child. no
.teachers keep asking for more with no difference in teaching just more money and if they dont get it what happens they strike .}well thats real nice on kids education is it not boo hoo you \claim{TAKE YOUR KIDS PRIVATE IF YOU CARE AS I DID} \newline

\noindent \textbf{Claim} $\bullet$ ``TAKE YOUR KIDS PRIVATE IF YOU CARE AS I DID'' \textbf{Premise} \emph{Teachers in public just want more money but it does not help the kids education} ``what about all the money we do send to our schools .does it help our child. no .teachers keep asking for more with no difference in teaching just more money and if they dont get it what happens they strike .''
\end{quote}
\end{example}

\begin{example}
This argument has been annotated as completely in the pathos dimension by only appealing to emotions (``send children to a pig farm'').

\begin{quote}
\textbf{Doc\#2581 (comment, public-private-schools)}
\appealtoemotion{Absolutely stupid person, why do we have children? To send to an immoral government run pig farm? No but to give to our children all the best that we as parents can!} \newline

\noindent \textbf{Claim} $\bullet$ \emph{Implicit: Against public schools} \textbf{Appeal to emotion} $\bullet$ ``Absolutely stupid person, why do we have children? To send to an immoral government run pig farm? No but to give to our children all the best that we as parents can!''
\end{quote}
\end{example}

\clearpage

\starttwocolumn

\begin{acknowledgments}
This work has been supported by the Volkswagen Foundation as part of the Lich\-ten\-berg-Professorship Program under grant N\textsuperscript{\underline{o}} I/82806, the German Institute for Educational Research (DIPF), and the German Research Foundation via the German-Israeli Project Cooperation (DIP, grant DA 1600/1-1). Computational resources were provided by the MetaCentrum under the program LM2010005 and the CERIT-SC under the program Centre CERIT Scientific Cloud, part of the Operational Program Research and Development for Innovations, Reg. no. CZ.1.05/3.2.00/08.0144. We would like to thank the anonymous reviewers for their valuable feedback and Judith Eckle-Kohler, Christian Stab, Emily Jamison, and Miloslav Konopik for their comments.
\end{acknowledgments}

\bibliographystyle{fullname}
\bibliography{bibliography}

\end{document}